\documentclass{article}

    \usepackage[final]{neurips_2022}

\usepackage[utf8]{inputenc} % allow utf-8 input
\usepackage[T1]{fontenc}    % use 8-bit T1 fonts
\usepackage{hyperref}       % hyperlinks
\usepackage{url}            % simple URL typesetting
\usepackage{booktabs}       % professional-quality tables
\usepackage{amsfonts}       % blackboard math symbols
\usepackage{nicefrac}       % compact symbols for 1/2, etc.
\usepackage{microtype}      % microtypography
\usepackage{xcolor}         % colors

\usepackage{graphicx}
\usepackage{amsmath}
\usepackage{amssymb}
\usepackage{booktabs}
\usepackage{mathtools}
\usepackage[symbol*]{footmisc}
\usepackage[export]{adjustbox}
\usepackage{multirow}
\usepackage[ruled,vlined]{algorithm2e}
\usepackage[inline]{enumitem}
\usepackage{nicefrac}

\usepackage{ifthen}
\usepackage[normalem]{ulem}
\usepackage{xcolor}
\usepackage[toc,page,header]{appendix}
\usepackage{minitoc}
\newtheorem{theorem}{Theorem}[section]
\newcommand{\thmref}[1]{Theorem~\ref{#1}}
\newcommand{\figref}[1]{Figure~\ref{#1}}
\renewcommand{\eqref}[1]{Eq.~(\ref{#1})}
\newcommand{\secref}[1]{Section~\ref{#1}}
\newcommand{\subsecref}[1]{Subsection~\ref{#1}}

\newcommand{\stam}[1]{}

\usepackage{mathtools}

\newcommand{\bx}{\mathbf{x}}

\newcommand{\btheta}{{\boldsymbol{\theta}}}

\newcommand{\cl}{{\cal L}}

\usepackage{bbm}

\newcommand{\reals}{{\mathbb R}}

\newcommand{\norm}[1]{\left\|#1\right\|}
\newcommand{\figtwod}{width=100pt}

\title{{Reconstructing Training Data from Trained\\ Neural Networks }}

\author{%
    Niv Haim\thanks{Equal contribution, alphabetically ordered} \\
    Weizmann Institute of Science \\
    \texttt{niv.haim@weizmann.ac.il} \\
    \And
    Gal Vardi$^*$\thanks{Work done while the author was at the Weizmann Institute of Science} \\
    TTI-Chicago and Hebrew University \\
    \texttt{galvardi@ttic.edu} \\
    \And
    Gilad Yehudai$^*$ \\
    Weizmann Institute of Science \\
    \texttt{gilad.yehudai@weizmann.ac.il} \\
    \And
    Ohad Shamir \\
    Weizmann Institute of Science \\
    \texttt{ohad.shamir@weizmann.ac.il} \\
    \And
    Michal Irani \\
    Weizmann Institute of Science \\
    \texttt{michal.irani@weizmann.ac.il} \\
    \vspace{4pt}
}

\begin{document}

\doparttoc % Tell to minitoc to generate a toc for the parts
\faketableofcontents % Run a fake tableofcontents command for the partocs
% \part{} % Start the document part
% \parttoc % Insert the document TOC

\maketitle

\begin{abstract}

\vspace{-60pt}Project page: \url{https://giladude1.github.io/reconstruction} \\
\vspace{32pt}

Understanding to what extent neural networks memorize training data is an intriguing question with practical and theoretical implications. 
In this paper we show that in some cases a significant fraction of the training data can in fact be reconstructed from the parameters of a trained neural network classifier.
We propose a novel reconstruction scheme that stems from recent theoretical results about the implicit bias in training neural networks with gradient-based methods.
To the best of our knowledge, our results are the first to show that reconstructing a large portion of the actual training samples from a trained neural network classifier is generally possible.
This has negative implications on privacy, as it can be used as an attack for revealing sensitive training data. 
We demonstrate our method for binary MLP classifiers on a few standard computer vision datasets.
\end{abstract}

\section{Introduction}

It is commonly believed that neural networks memorize the training data, even when they are able to generalize well to unseen test data (e.g.,  \citep{zhang2021understanding,feldman2020does}). Exploring this memorization phenomenon is of great importance both practically and theoretically.
Indeed, it has implications on our understanding of generalization in deep learning, on the hidden representations learnt by neural networks, and on the extent to which they are vulnerable to privacy attacks.

\begin{figure}
    \centering
    % \advance\leftskip-2cm
    \begin{tabular}{c}
        (a) Top 24 images reconstructed from a binary classifier trained
    on 50 CIFAR10 images \\
         \includegraphics[width=\textwidth]{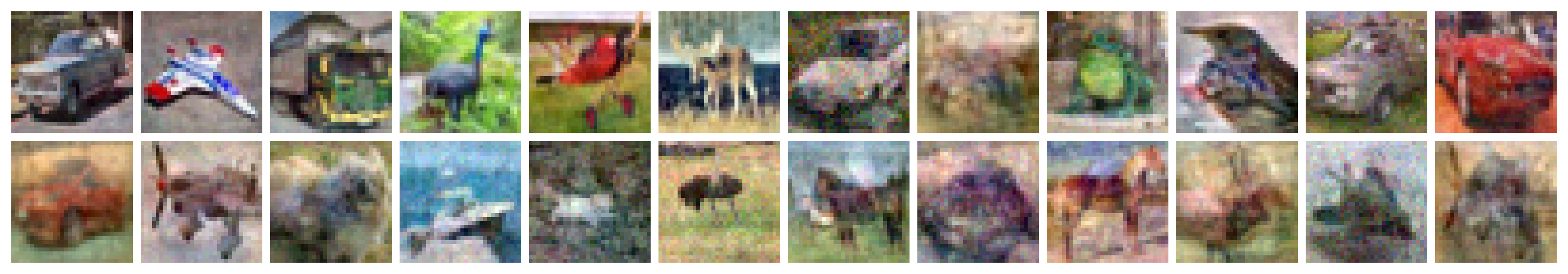} \\
         (b) Their corresponding nearest neighbours from the training-set of the model \\
      \includegraphics[width=\textwidth]{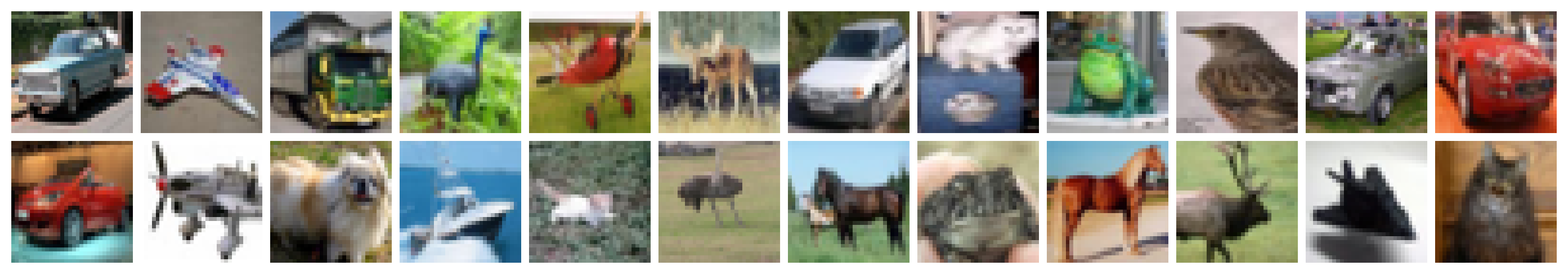} \\
    \end{tabular}
    \caption{Reconstruction of training images from a pretrained binary classifier, trained on 50 CIFAR10 images. The two classes are ``animals'' and ``vehicles''. We calculate the nearest neighbor using the SSIM metric.\vspace{-12pt}}
    \label{fig:teaser}
\end{figure}

A fundamental question for understanding memorization is:
\begin{quote}
\emph{Are the specific training samples encoded in the parameters of a trained classifier? Can they be recovered from the network parameters?}
\end{quote}
In this work, we study this question, and devise a novel scheme which allows us to reconstruct a significant portion of the training data 
from the parameters of a trained neural network alone, without having any additional information on the data.
Thus, we provide a proof-of-concept that the learning process can sometimes be reversed: That is, instead of learning a 
model given a training dataset, it is possible to find the training data given a trained model. In \figref{fig:teaser} we show how our approach reconstructs images from the CIFAR10 dataset, given a simple trained binary classifier.

Many works try to ``crack'' neural networks by analyzing and visualizing either their learnt parameters or representations  \citep{erhan2009visualizing,mahendran2015understanding,olah2017feature,olah2020zoom}. This is usually done by ``inverting'' the model, namely finding inputs that are strongly correlated with the model's activations \citep{mordvintsev2015inceptionism,yin2020dreaming,fredrikson2015model}. Unsurprisingly, the results are semantically correlated with the training dataset. However, one rarely sees an exact version of a training sample.

Our results have potential negative implications on privacy in deep learning.
Our scheme can be viewed as a \emph{training-data reconstruction attack}, since an adversary might recover sensitive training data. For example, if a medical device includes a model trained on sensitive medical records, an adversary might reconstruct this data and thus violate the privacy of the patients. Privacy attacks in deep learning have been widely studied in recent years 
(cf. \cite{liu2021machine}),
but as far as we are aware, the known attacks cannot reconstruct portions of the training data from a trained model.

Our approach relies on theoretical results about the implicit bias in training neural networks with gradient-based methods. The implicit bias has been studied extensively in recent years with the motivation of explaining generalization in deep learning (see \secref{sec:related}).
We use results by \cite{lyu2019gradient,ji2020directional}, which establish that, under some technical assumptions, if we train a neural network with the binary cross entropy loss, its parameters will converge to a stationary point of a certain margin-maximization problem. This result implies that the parameters of the trained network satisfy a set of equations w.r.t. the training dataset. In our approach, given a trained network, we find a dataset that solves this set of equations w.r.t. the trained parameters. 

\vspace{-6pt}
\paragraph{Our Contributions} We show that large portions of the training samples are encoded in the parameters of a trained classifier. We also provide a practical scheme to decode the training samples, without any assumptions on the data. As far as we know, this is the first work that shows that reconstruction of actual training samples from a trained neural network classifier is possible.

\vspace{-6pt}
\section{Related Work} \label{sec:related}

\paragraph{Understanding and Visualizing what is learnt by Neural Networks.}
The most common approach for analysing what is learnt by a neural network is by searching inputs that maximize the 
class output or the activations of neurons in intermediate layers~\citep{erhan2009visualizing,olah2020zoom}. Oftentimes this is done via optimization with respect to the model input. 
Optimizing without any prior on the input usually results in noise inputs. Therefore, most approaches incorporate priors such as smoothness regularization or the use of pre-trained image generators \citep{mahendran2015understanding,yosinski2015understanding,mordvintsev2015inceptionism,nguyen2016synthesizing,nguyen2016multifaceted,nguyen2017plug} (see \cite{olah2017feature} for a comprehensive summary). 
Optimization w.r.t. the input may also result in adversarial examples \citep{szegedy2013intriguing,goodfellow2014explaining}. Recently, \citep{tsipras2018robustness,engstrom2019adversarial} showed that classifiers trained to be robust to adversarial examples tend to learn representations that are more aligned with human vision. This was later utilized by \citep{santurkar2019image,mejia2019robust} to generate class-conditional images from a trained classifier.
While all those approaches indicate that, unsurprisingly, the learnt representations are strongly correlated with the datasets on which the model was trained, none of them demonstrate the reconstruction of exact training samples from the trained models.

\paragraph{Privacy Attacks in Deep Learning.}

Many methods deal with extracting sensitive information from trained models. Perhaps the closest to our approach is
\emph{model-inversion} that aims to reconstruct class representatives from the training data of a trained model \citep{fredrikson2015model,he2019model,yang2019neural,yin2020dreaming}. It is important to note that the reconstructed images, albeit semantically similar to some input images, are still not actual samples from the training set. \cite{carlini2021extracting,carlini2019secret} demonstrated reconstruction of training data from generative language models. By completing sentences, they reveal sensitive information from the training data. We note that this approach is specific to generative language models, while our approach considers classifiers and is less data specific.
\emph{Membership-inference} attacks \citep{shokri2017membership} aim to determine whether a given data point was used to train the model or not. For these methods to work, the adversary must be able to guess a specific input, whereas our approach does not assume such ability.
Lastly, avoiding leakage of sensitive information on the training dataset is the motivation behind \emph{differential privacy} in machine learning, which has been extensively studied \citep{abadi2016deep,dwork2006calibrating,chaudhuri2011differentially}. For an elaborated discussion 
on the relation of these approaches to ours see Appendix~\ref{app:more related}.

\paragraph{Implicit Bias.}

In overparameterized neural networks one might expect overfitting to occur, but it seems that gradient-based methods are biased towards networks that generalize well \citep{zhang2021understanding,neyshabur2017exploring}. Mathematically characterizing this implicit bias is a major problem in the theory of deep learning.
Our approach is based on a characterization of the implicit bias of gradient flow in homogeneous neural networks due to \cite{lyu2019gradient} and \cite{ji2020directional} (see \secref{sec:back and met} for details). The implicit bias of gradient-based methods in neural networks was extensively studied in recent years both for classification tasks (e.g., \cite{soudry2018implicit,gunasekar2018bimplicit,ji2018gradient,nacson2019lexicographic,vardi2021margin,chizat2020implicit,gunasekar2018characterizing,moroshko2020implicit}) and regression tasks (e.g., \cite{gunasekar2018implicit,arora2019implicit,azulay2021implicit,yun2020unifying,woodworth2020kernel,razin2020implicit,li2020towards,vardi2021implicit,timor2022implicit}). See \cite{vardi2022implicit} for a survey.

\section{Background and Reconstruction Scheme} \label{sec:back and met}
In this section we present our training data reconstruction scheme, as well as provide a brief overview on the theoretical results about implicit bias, which motivate our approach.

\subsection{On the Implicit Bias of Neural Networks}

Let $S = \{(\bx_i,y_i)\}_{i=1}^n \subseteq \reals^d \times \{-1,1\}$ be a binary classification training dataset. Let $\Phi(\btheta;\cdot):\reals^d \to \reals$ be a neural network parameterized by $\btheta \in \reals^p$. 
For a loss function $\ell:\reals \to \reals$ the empirical loss of $\Phi(\btheta; \cdot)$ on the dataset $S$ is 
$\cl(\btheta) := \sum_{i=1}^n \ell(y_i \Phi(\btheta; \bx_i))$.
We focus on 
the \emph{logistic loss} (a.k.a. \emph{binary cross entropy}), namely, $\ell(q) = \log(1+e^{-q})$. 

Our approach is based on 
\thmref{thm:known KKT} below, 
which holds for \emph{gradient flow} (i.e., gradient descent with an infinitesimally small step size). 
Before stating the theorem, we need the following definitions:
(1) We say that 
gradient flow
{\em converges in direction} to $\tilde{\btheta}$ if $\lim_{t \to \infty}\frac{\btheta(t)}{\norm{\btheta(t)}} = \frac{\tilde{\btheta}}{\norm{\tilde{\btheta}}}$, where $\btheta(t)$ is the parameter vector at time $t$;
(2) We say that a network $\Phi$ is \emph{homogeneous} w.r.t. the parameters $\btheta$ if there exists $L>0$ such that for every $\alpha>0$ and $\btheta,\bx$ we have $\Phi(\alpha \btheta; \bx) = \alpha^L \Phi(\btheta; \bx)$. 
Thus, scaling the parameters by any factor $\alpha>0$ scales the outputs by $\alpha^L$. 
We note that essentially any fully-connected or convolutional neural network with ReLU activations is homogeneous w.r.t. the parameters $\btheta$ if it does not have any skip-connections (i.e., residual connections) or bias terms, except possibly for the first layer.

\begin{theorem}[Paraphrased from \cite{lyu2019gradient,ji2020directional}]
\label{thm:known KKT}
	Let $\Phi(\btheta;\cdot)$ be a homogeneous 
	ReLU neural network. Consider minimizing 
	the logistic loss over a binary classification dataset $ \{(\bx_i,y_i)\}_{i=1}^n$ using gradient flow. Assume that there exists time $t_0$ such that $\cl(\btheta(t_0))<1$\footnote{This ensures that $\ell(y_i \Phi(\btheta(t_0); \bx_i)) <1$ for all $i$, i.e. at some time $\Phi$ classifies every sample correctly.}.
	Then, gradient flow converges in direction to a first order stationary point (KKT point) of the following maximum-margin problem:
\begin{equation}
\label{eq:optimization problem}
	\min_{\btheta'} \frac{1}{2} \norm{\btheta'}^2 \;\;\;\; \text{s.t. } \;\;\; \forall i \in [n] \;\; y_i \Phi(\btheta'; \bx_i) \geq 1~.
\end{equation}
Moreover, $\cl(\btheta(t)) \to 0$ 
as $t \to \infty$.
\end{theorem}

The above theorem guarantees directional convergence to a first order stationary point (of the optimization problem ~(\ref{eq:optimization problem})), which is also called \emph{Karush–Kuhn–Tucker point}, or \emph{KKT point} for short. 
The KKT approach allows inequality constraints, and is a generalization of the method of \emph{Lagrange multipliers}, which allows only equality constraints.

The great virtue of \thmref{thm:known KKT} is that it characterizes the \emph{implicit bias} of gradient flow with the logistic loss for homogeneous networks. Namely, even though there are many possible directions of $\frac{\btheta}{\norm{\btheta}}$ that classify the dataset correctly, gradient flow converges only to directions that are KKT points of Problem~(\ref{eq:optimization problem}). In particular, if the trajectory $\btheta(t)$  
of gradient flow under the regime of \thmref{thm:known KKT} converges in direction to a KKT point $\tilde{\btheta}$, then 
we have the following:
There exist $\lambda_1,\ldots,\lambda_n\in\reals$ such that 
\begin{align} 
    &\tilde{\btheta} = \sum_{i=1}^n \lambda_i y_i \nabla_\btheta  \Phi(\tilde{\btheta}; \bx_i)~ &\text{(stationarity)}\label{eq:stationary}\\
    & \forall i \in [n], \;\; y_i \Phi(\tilde{\btheta}; \bx_i) \geq 1 &\text{(primal feasibility)}\label{eq:prim feas} \\
    &\lambda_1,\ldots,\lambda_n \geq 0 &\text{(dual feasibility)}\label{eq:dual feas}\\
    & \forall i \in [n],~~ \lambda_i= 0 ~ \text{if}~  y_i \Phi(\tilde{\btheta}; \bx_i) \neq 1 & \text{(complementary slackness)}\label{eq:comp slack}
\end{align}
Our main insight is based on \eqref{eq:stationary}, which implies that the parameters $\tilde{\btheta}$ are a linear combinations of the derivatives of the network at the training data points. 
We say that a data point $\bx_i$ is \emph{on the margin} if $y_i \Phi(\tilde{\btheta}; \bx_i) = 1$ (i.e. $|\Phi(\tilde{\btheta}; \bx_i)|=1$) .
Note that \eqref{eq:comp slack} implies that only samples which are on the margin 
affect \eqref{eq:stationary},
since samples not on the margin have a coefficient $\lambda_i=0$.

\subsection{Dataset Reconstruction}
\label{subsec:dataset_reconstruction}
Suppose we are given a trained neural network with parameters $\btheta$, and our goal is to reconstruct the dataset that the network was trained on. 
%Using 
Although \thmref{thm:known KKT} holds asymptotically as the time $t$ tends to infinity, it suggests that also after training for a finite number of iterations the parameters of the network might approximately satisfy \eqref{eq:stationary}, and the coefficients $\lambda_i$ satisfy \eqref{eq:dual feas}. 
Since $n$ is unknown (and so is the number of samples on the margin) we set $m \geq 2n$ 
which represents the number of samples we want to reconstruct (thus, we only need to upper bound $n$), and 
fix 
$y_i = 1$ for $i = 1,\ldots,m/2$ and $y_i=-1$ for $i = m/2+1,\ldots,m$.
We define the following losses:
\begin{align}
    &L_{\text{stationary}}(\bx_1,\dots,\bx_m,\lambda_1,\dots,\lambda_m) =  \left\| \btheta - \sum_{i=1}^m \lambda_i y_i \nabla_\btheta  \Phi(\btheta; \bx_i) \right\|_2^2\label{eq:kkt loss}\\
    &L_\lambda(\lambda_1,\dots,\lambda_m) = \sum_{i=1}^m \max\{-\lambda_i, 0\}\label{eq:lambda loss}
\end{align}
Note that the unknown parameters are the $\bx_i$'s and $\lambda_i$'s, and that $\btheta$ and the $y_i$'s are given.
The loss $L_{\text{stationary}}$ represents the stationarity condition that the parameters of the network satisfy, and $L_\lambda$ represents the dual feasibility condition.
We additionally define $L_{\text{prior}}$ which represents some prior knowledge we might have about the dataset. For example, if we know that the dataset contains images, prior knowledge would be that each input coordinate (i.e. each pixel) is between $0$ and $1$. Given no prior knowledge on the data, we can define $L_{\text{prior}}\equiv 0$. Finally, we define the reconstruction loss as:
\begin{equation}\label{eq:extraction loss}
    L_{\text{reconstruct}}(\left\{ \bx_i \right\}_{i=1}^m,\left\{ \lambda_i \right\}_{i=1}^m) = \alpha_1 L_{\text{stationary}} + \alpha_2 L_\lambda + \alpha_3 L_{\text{prior}}
\end{equation}
where $\alpha_1,\alpha_2,\alpha_3\in\reals$ are tunable hyperparameters of the different losses. To reconstruct the dataset, we can use any nonconvex optimization method (e.g. SGD) to find the $\bx_1,\dots,\bx_m,\lambda_1,\dots,\lambda_m$ which minimize \eqref{eq:extraction loss}. We note that the $\lambda_i$'s are not part of the training data, but finding them is necessary in order to solve this optimization problem. Finally, we emphasize that there are many other possible options to formulate the KKT conditions Eq.~(\ref{eq:stationary})-(\ref{eq:comp slack}) as an unconstrained optimization problem. However, this simple choice seemed to work quite well in practice.

We note that if there exist $\{\bx_i\}_{i=1}^n$ and $\{\lambda_i\}_{i=1}^n$ which satisfy the KKT conditions, then there are $\{\bx_i\}_{i=1}^m$ and $\{\lambda_i\}_{i=1}^m$ which achieve zero loss in \eqref{eq:extraction loss}. Indeed, such a solution can be obtained by adding to $\{\bx_i\}_{i=1}^n$ additional points $\bx_j$ with $\lambda_j=0$, or by duplicating some points in $\{\bx_i\}_{i=1}^n$ and modifying the $\lambda$'s accordingly. Also, note that since we choose $m \geq 2n$, then we set at least $n$ labels $y_i$ to $1$ and at least $n$ labels to $-1$. Hence, there is a solution to \eqref{eq:extraction loss} even though we do not know the real distribution of labels in the actual training data.  

We cannot simply use \eqref{eq:prim feas} and~(\ref{eq:comp slack}) in our reconstruction scheme, because they contain the constant "$1$" which corresponds to the margin (i.e., $\min_i \vert \Phi(\tilde{\btheta};\bx_i) \vert$). Namely, we only converge \emph{in direction} to a point $\tilde{\btheta}$ that attains margin $1$, but in practice we 
% converge to 
approach
some point $\btheta$ which attains an unknown margin $\gamma$ (i.e., $\min_i \vert \Phi(\btheta;\bx_i) \vert=\gamma$), and we do not know in advance how to normalize it to attain a margin of exactly $1$.  On the other hand, \eqref{eq:stationary} and~(\ref{eq:dual feas}) hold not only for $\tilde{\btheta}$ but also for any $\btheta$ that points at the direction of $\tilde{\btheta}$, and therefore in our loss 
in \eqref{eq:extraction loss}
we rely only on these conditions.

Intuitively, a reason to believe that there is enough information in \eqref{eq:stationary} to reconstruct the data, is the following observation:
\eqref{eq:stationary} represents a set of $p$ equations with $O(nd)$ unknown variables, where $p$ is the number of parameters in the network. In practice, neural networks are often highly overparameterized (i.e., $p > nd$), suggesting more equations than variables.

Finally, since by \eqref{eq:comp slack} we have $\lambda_i=0$ for every $\bx_i$ that is not on the margin, then \eqref{eq:stationary} implies that $\tilde{\btheta}$ is determined only by the gradients w.r.t. the data points that are on the margin. 
Hence, we can only expect to reconstruct training samples that are on the margin (see also Subsection \ref{subsec:margin}).

\section{A Simple Experiment in Two Dimensions}\label{sec:2d}

\begin{figure}[ht]
    \centering
    \advance\leftskip-.6cm
    \begin{tabular}{cccc}
        \expandafter\includegraphics\expandafter[\figtwod]{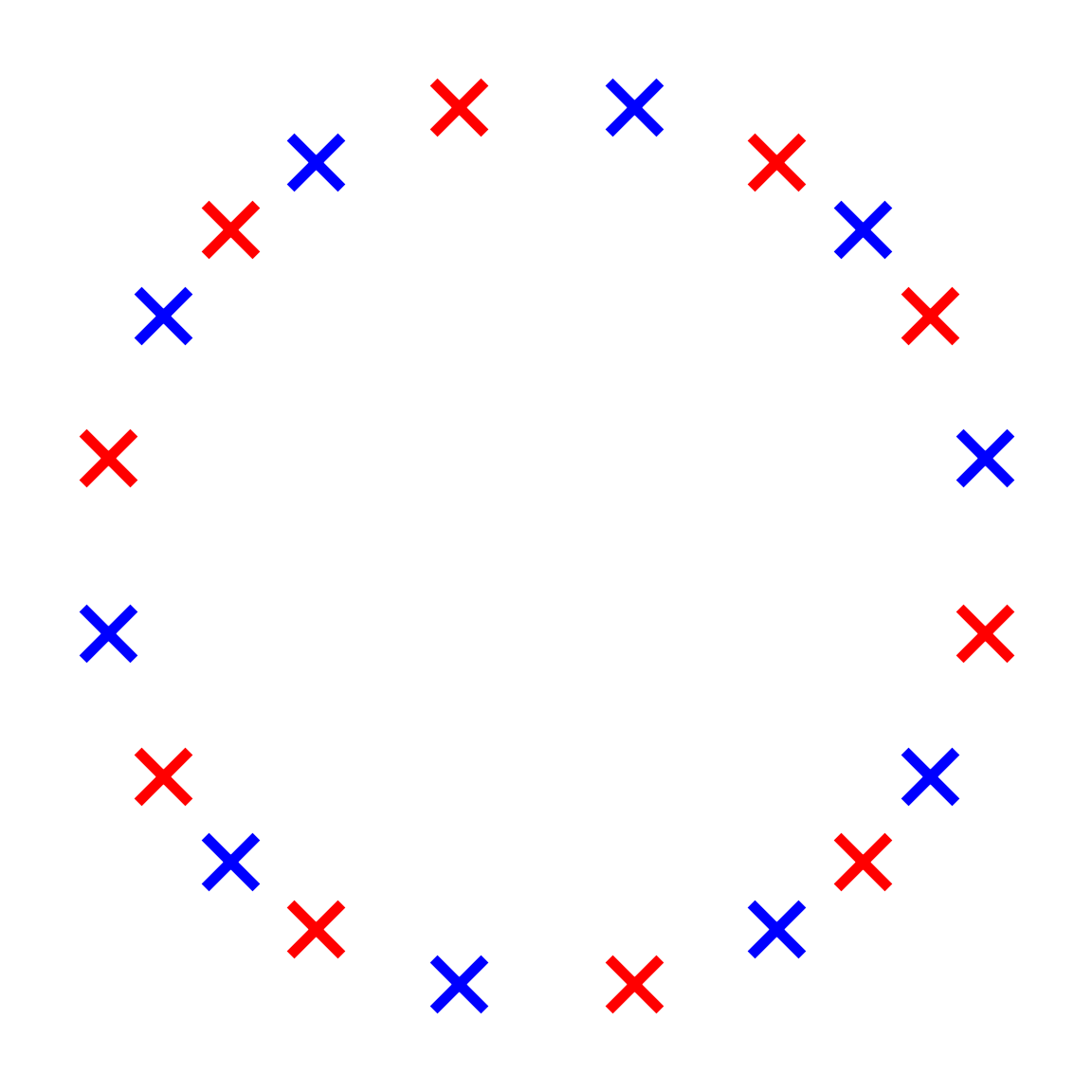} &
        \expandafter\includegraphics\expandafter[\figtwod]{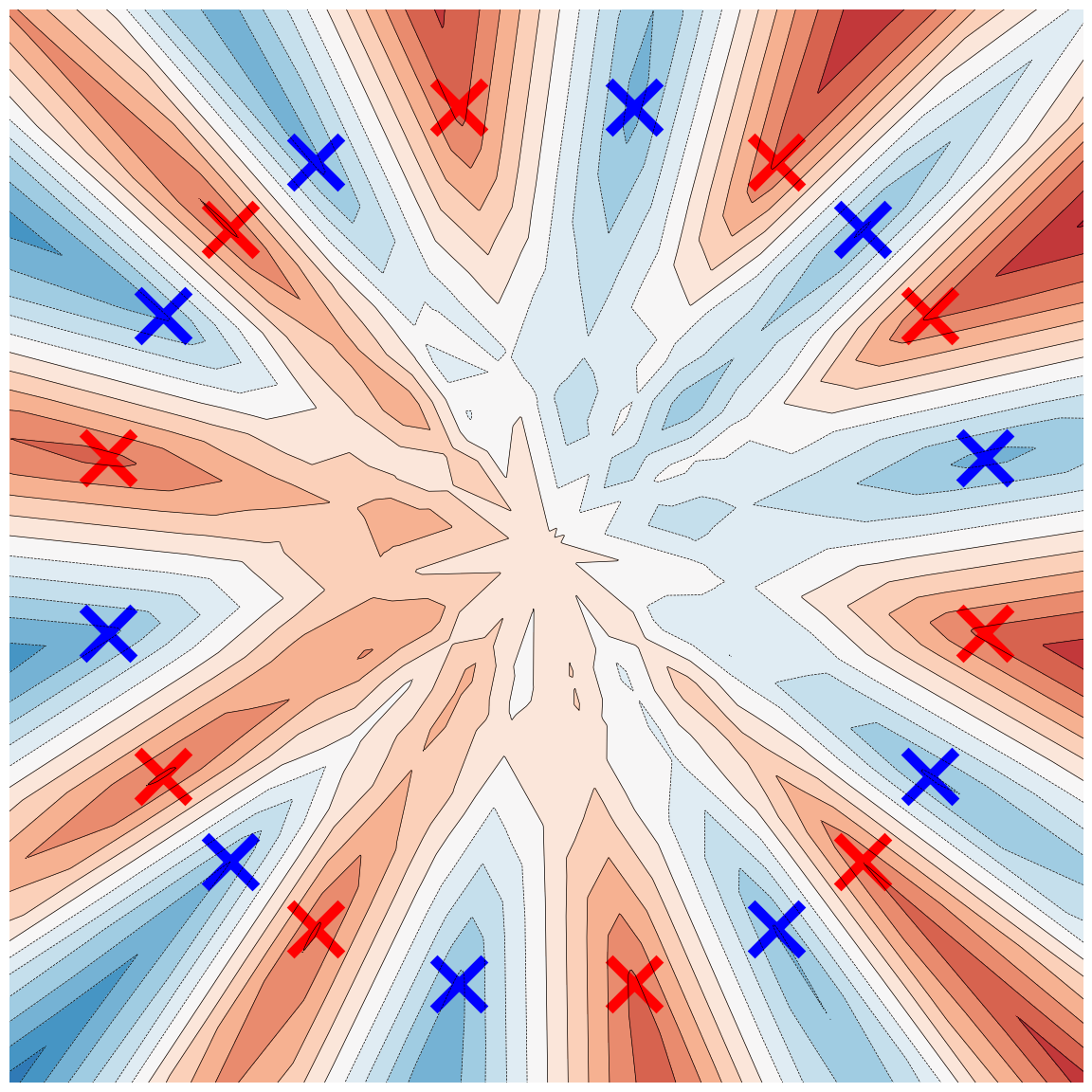} &
        \expandafter\includegraphics\expandafter[\figtwod]{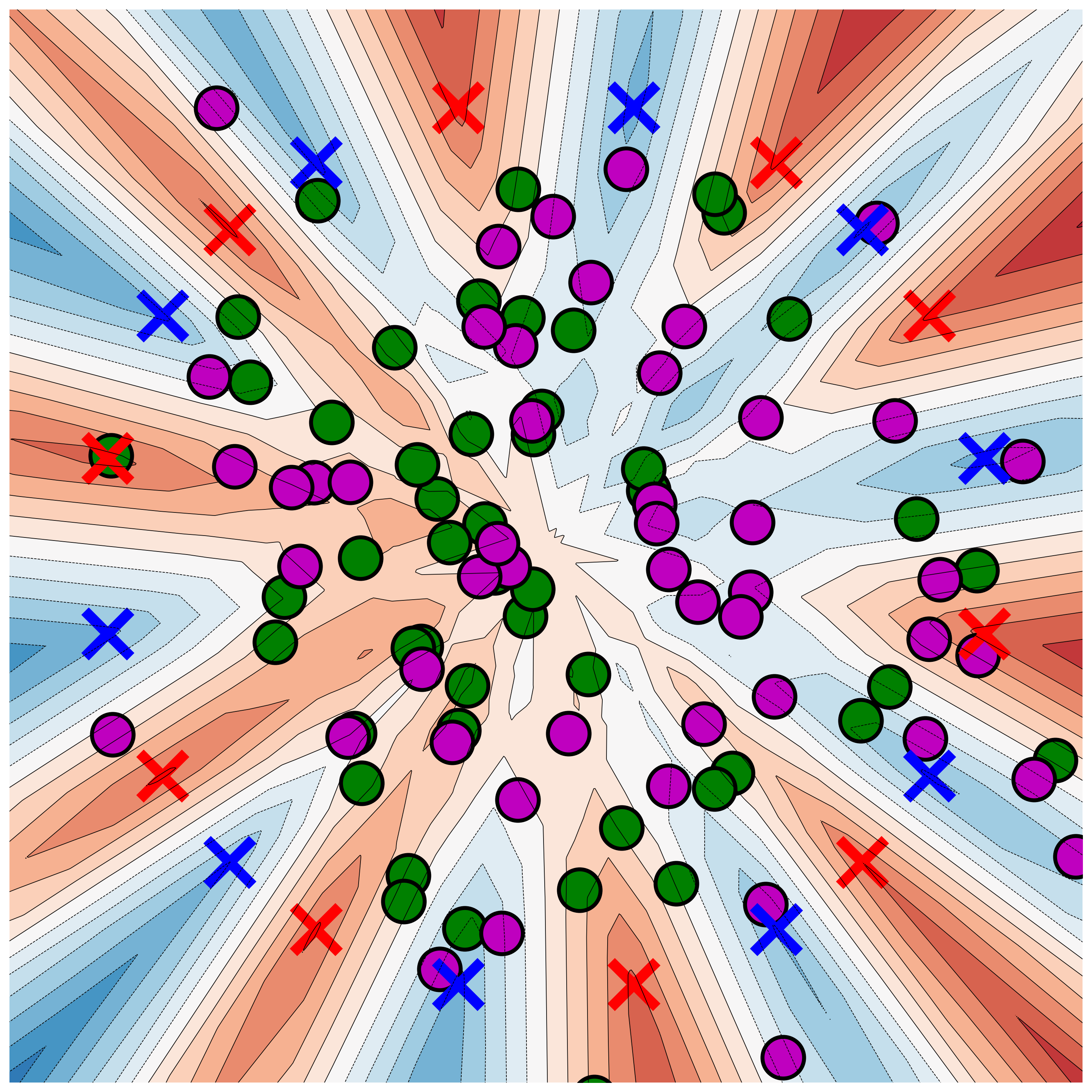}& \\
        (a) Training Set & (b) Model Landscape & (c) Initialization & \\
        \expandafter\includegraphics\expandafter[\figtwod]{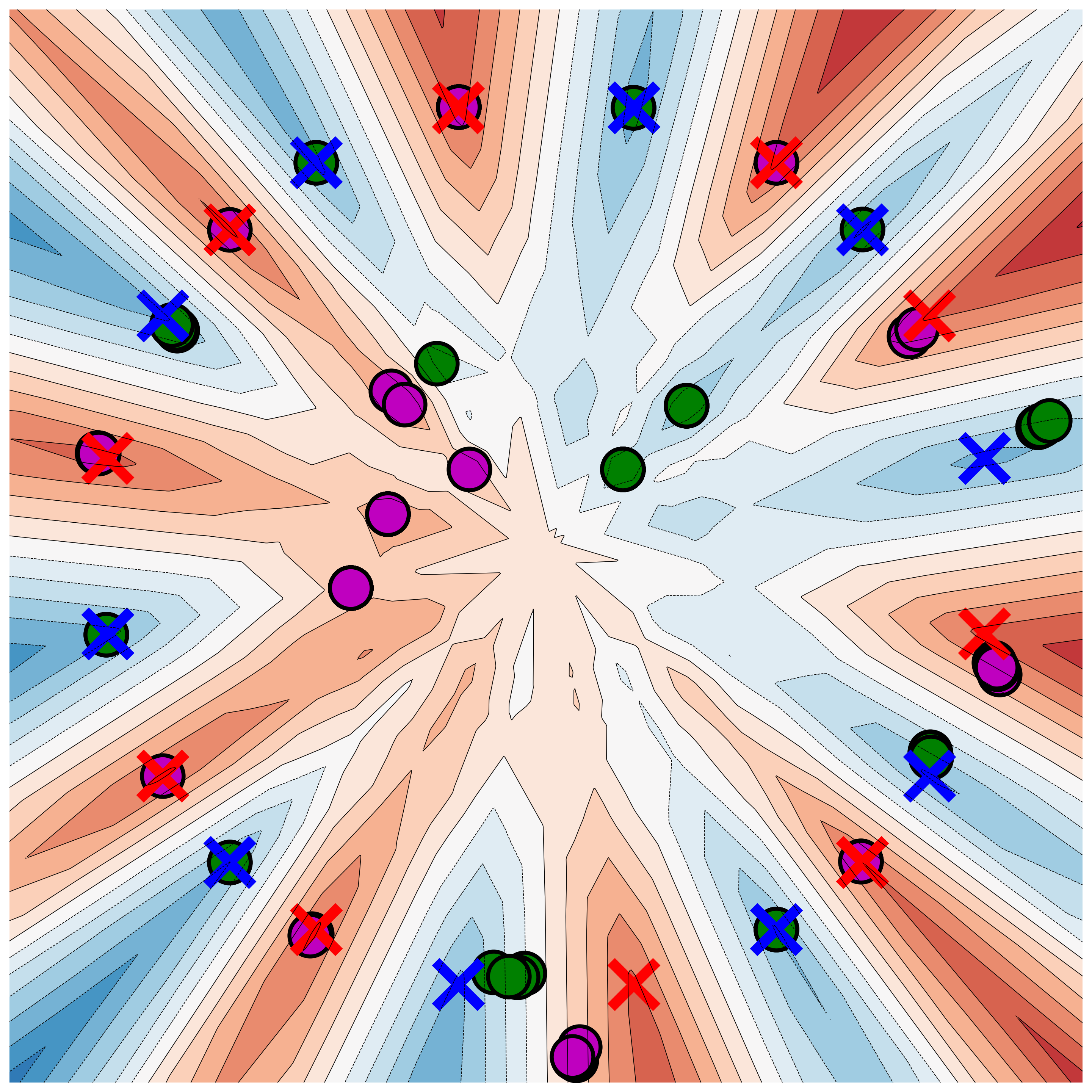} &
        \expandafter\includegraphics\expandafter[\figtwod]{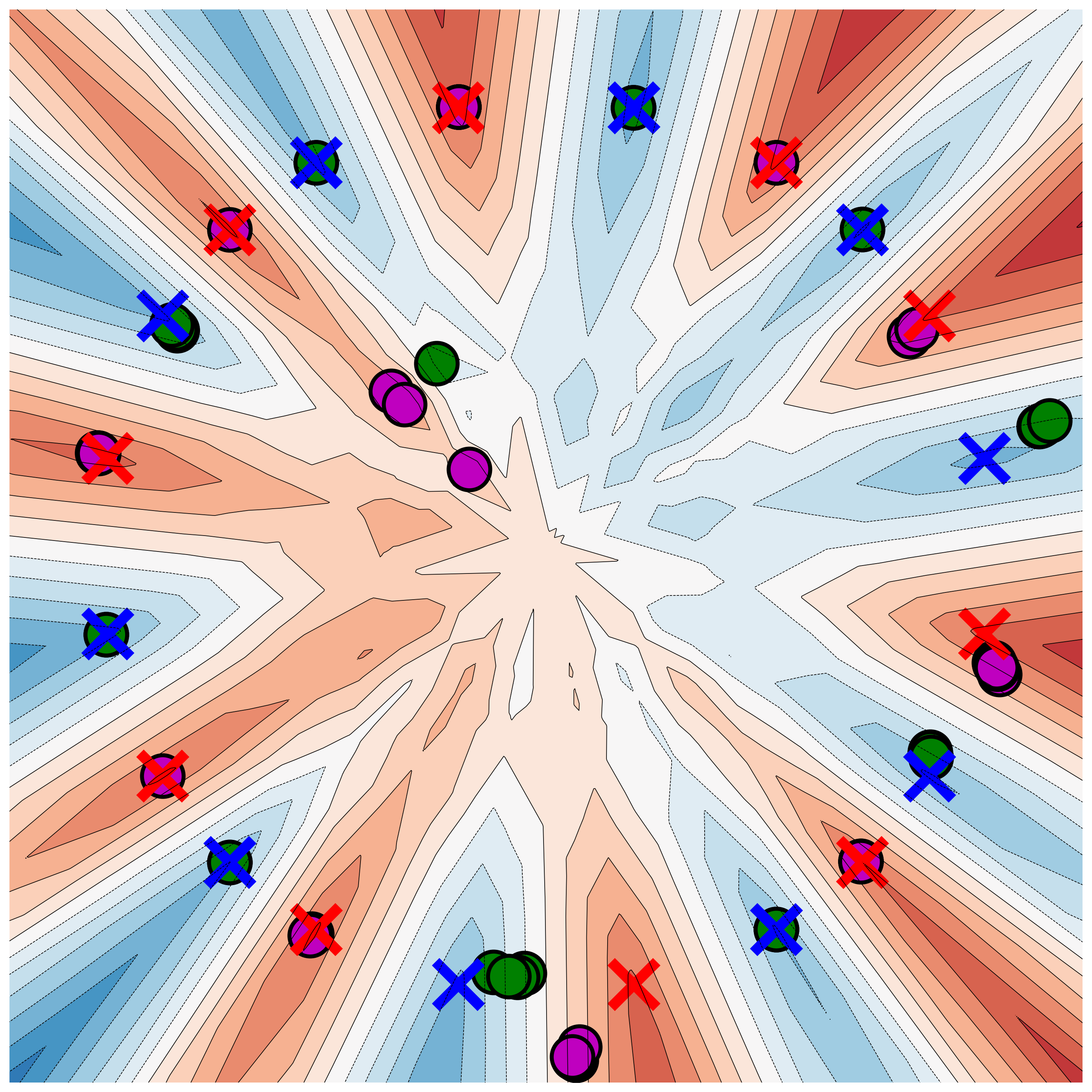} &
        \expandafter\includegraphics\expandafter[\figtwod]{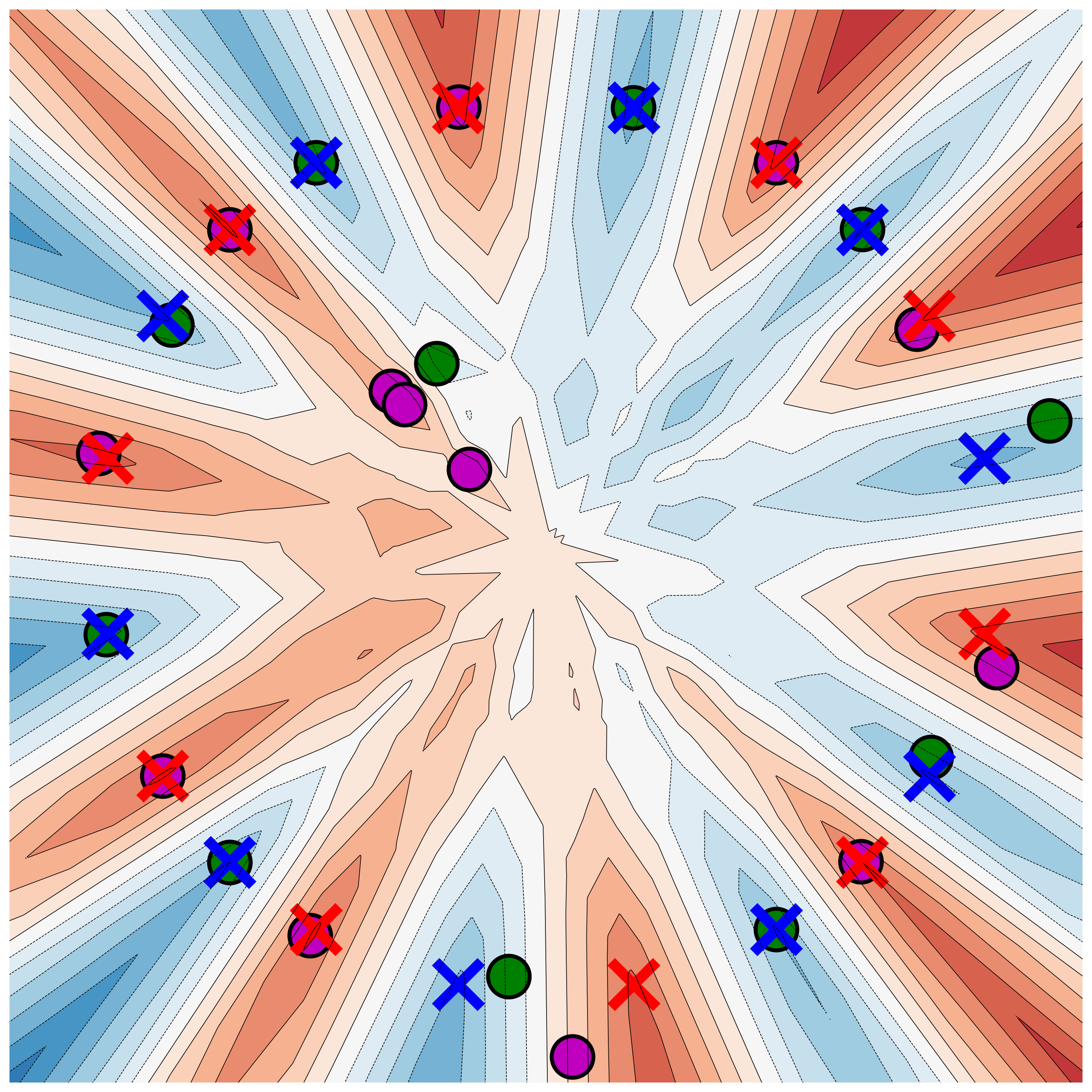} &\\
        (d) Reconstruction & (e) Removing Small  $\lambda$'s & (f) Removing Duplicates &
    \end{tabular}
    \caption{Exemplifying our reconstruction scheme on a simple 2D dataset (see text for explanation).}
    \label{fig:2d circle}
\end{figure}

In this section we exemplify our dataset reconstruction scheme on a toy example of $2$-dimensional data, i.e. we consider $(\bx,y)\in \reals^2 \times \{\pm 1\}$. We set $n=20$ training samples on the unit circle, with alternating labels. For a visualization of the dataset see \figref{fig:2d circle}a, blue and red "$\times$" represent the two classes. We trained a $3$-layer model with $1000$ neurons in each layer on this dataset. The model learns to correctly classify the training set. In \figref{fig:2d circle}b, we visualize the output of the model as a function of its input. Blue and red regions correspond to smaller and larger outputs of the model, respectively.

We now demonstrate our reconstruction scheme. We first randomly initialize $m=100$ points in $\reals^2$, and assign $50$ points to each class. This is depicted in \figref{fig:2d circle}c, where green points correspond to the blue class, and magenta points correspond to the red class.  Next, we optimize the loss in \eqref{eq:extraction loss}, with $L_{\text{prior}}\equiv 0$. The results of our reconstruction scheme are in \figref{fig:2d circle}d. Note that our approach reconstructed all the input samples, up to some noise.

To further improve our reconstruction results, we remove some of the extra points which did not converge to a training sample. In \figref{fig:2d circle}e we removed points $\bx_i$ with corresponding $\lambda_i < 5$. According to \eqref{eq:stationary}, points with $\lambda_i = 0$ should not affect the parameters, hence their corresponding $\bx_i$ can take any value. In practice, it is sufficient to remove points with a small enough corresponding $\lambda_i$. Finally, to remove duplicates, we greedily remove points which are very close to other points. That is, we randomly order the points, and iteratively remove points that are at distance $<0.03$ from another point. The final reconstruction result is depicted in \figref{fig:2d circle}f.

\section{Results}\label{sec:results}

\begin{figure}[ht]
    \centering
    % \advance\leftskip-2cm
    \begin{tabular}{l}
    Top $45$ images reconstructed from a  model trained on CIFAR10 (rows $1,3,5$), and their corresponding \\ nearest-neighbors 
    from the training-set of the model (rows $2,4,6$) \\
    \includegraphics[width=\textwidth]{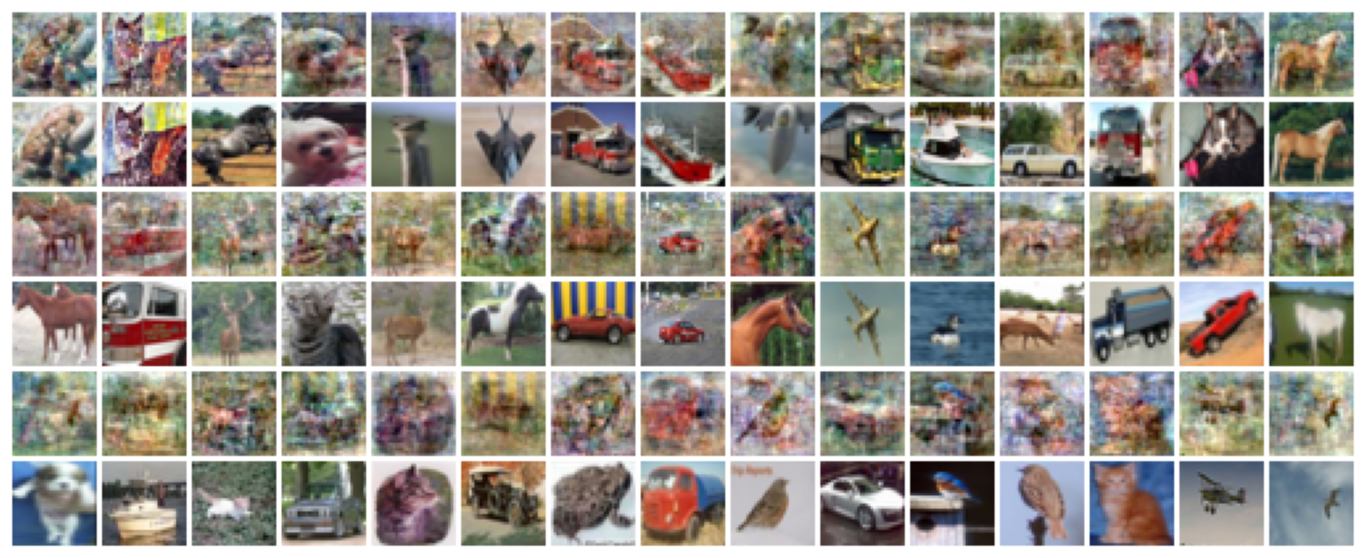} \\
     Top $45$ images reconstructed from a model trained on MNIST (rows $1,3,5$), and their corresponding \\ nearest-neighbors from the training-set of the model (rows $2,4,6$)  \\
     \includegraphics[width=\textwidth]{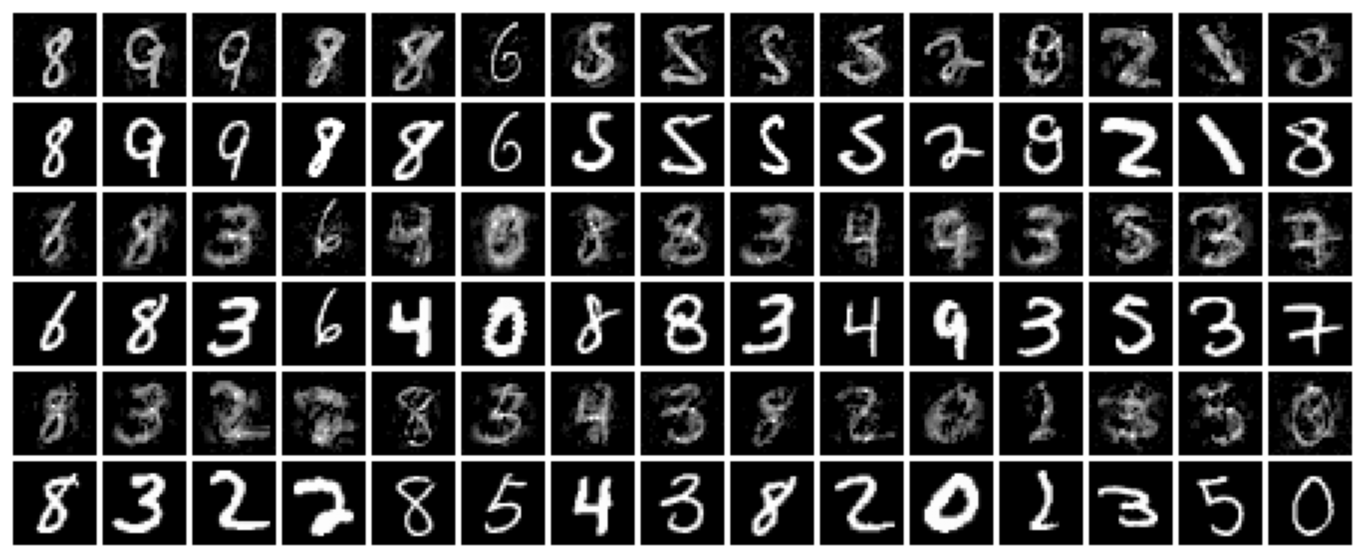} \\
    \end{tabular}
    \caption{Reconstructing training samples from two 
    binary classifiers -- one trained on $500$ images with labels animals/vehicles (CIFAR), and the other trained on $500$ odd/even digit images (MNIST). Train errors are zero, test accuracies are $88.0\% / 77.6\%$ for MNIST/CIFAR 
    }
    \label{fig:reconstruction}
\end{figure}

\subsection{Experimental Setup}
\paragraph{Datasets.} 
We conduct experiments on binary classification tasks where images are taken from the MNIST~\citep{lecun2010mnist} and CIFAR10~\citep{krizhevsky2009learning} datasets and the labels are set to odd vs. even digits (MNIST), and vehicles vs. animals\footnote{Automobile, Truck, Airplane, Ship vs. Bird, Horse, Cat, Dog, Deer, Frog.} (CIFAR10). We make sure that the class distribution in the training and test sets is balanced, and normalize the train and test sets by reducing the mean of the training set from both.

\paragraph{Training.} 
We consider MLP architectures. Unless stated otherwise, our models comprise of three fully-connected layers with dimensions $d$-$1000$-$1000$-$1$ (where $d$ is the dimension of the input) with ReLU activations. Biases are set to zero except for the first layer, to line up with the theoretical assumption of homogeneous models in \secref{sec:back and met}. 
The parameters are initialized using standard Kaiming He initialization~\citep{he2015delving} except for the weights of the first layer that are initialized to a Gaussian distribution with standard deviation $10^{-4}$ (see discussion in \subsecref{subsec:ts_reconstruct}). 
We train our models using full batch gradient descent for $10^6$ epochs with a learning rate of $0.01$. All models achieve zero training error (i.e., all the train samples are labeled correctly), and a training loss $<10^{-6}$. To compute the test accuracy, we use the original test sets of MNIST/CIFAR10 with $10000$/$8000$ images respectively, and labeled accordingly.

\subsection{Training Set Reconstruction}
\label{subsec:ts_reconstruct}
We minimize the loss defined in \eqref{eq:extraction loss} with $\alpha_1=1,~\alpha_2=5,~\alpha_3=1$.
We initialize $\bx_i \sim \mathcal{N}(0,\sigma_x I)$, where $\sigma_x$ is a hyperparameter, and $\lambda_i \sim \mathcal{U}[0,1]$. We set the number of reconstructed samples to $m=2n$ (where $n$ is the size of the original training set). 
Note that our loss contains the derivative of ReLU \eqref{eq:kkt loss}. This derivative is a step function, containing only flat regions which are hard to optimize.
We replace the derivative of the ReLU layer (backward function) with a sigmoid,
which is the derivative of softplus (a smooth version of ReLU).
We use the fact that our inputs are images to penalize values outside the range $[-1,1]$. To this end we set $L_{\text{prior}}(z) = \max\{z-1,0\} + \max\{-z-1,0\}$ for each pixel $z$, and average over all dimensions (pixels) in $\bx_i$. 
We optimize our loss for $100,000$ iterations using an SGD optimizer with momentum $0.9$. We conduct a total of $100$ runs using a random grid search on the hyperparameters (e.g. learning rate, $\sigma_x$. See Appendix~\ref{appen:implementation} for full details). This results in $100m$ ``reconstructed'' inputs.

While some $\bx_i$ end up converging to a training sample, some end as noise (similar phenomenon can be observed in 2D in \figref{fig:2d circle}d). To identify the reconstructions 
that are most similar to a training image we use the SSIM metric \citep{wang2004image}.

In \figref{fig:reconstruction} we show the best reconstruction results (in terms of SSIM) for models trained on $n$=$500$ samples from MNIST/CIFAR10 datasets (with test accuracy $88.0\%/77.6\%$ resp.). Note that the reconstructed images are very similar to the real input data, although a bit noisy. The source of this noise is not entirely clear. Possible reasons may be the complexity of the optimization problem, or the possibility that the trained model has not fully converged to the KKT point of Problem~(\ref{eq:optimization problem}).

We observed that small initializations significantly improve the quality of the reconstructed samples. 
We conjecture that small initialization causes faster convergence to the direction of the KKT point. This is also theoretically implied in \cite{moroshko2020implicit} (for certain linear models). Similarly, training for more epochs also improves the quality of the reconstruction. In Appendix~\ref{appen:results} we show results for reconstructions from networks trained with standard initialization or trained for much fewer epochs. During the training phase, we used full batch gradient descent, to remain as much aligned to the theoretical setting. In Appendix~\ref{appen:results} we show that our approach can reconstruct training data also from models trained with mini-batch SGD.

\subsection{Practice vs. Theory}
\label{subsec:margin}

\begin{figure}
    \centering
    \advance\leftskip-.7cm
    \begin{tabular}{c}
        Models with the same architecture ($1000$-$1000$) trained on different number of training samples ($n$)\vspace{6pt} \\
        \includegraphics[width=\textwidth]{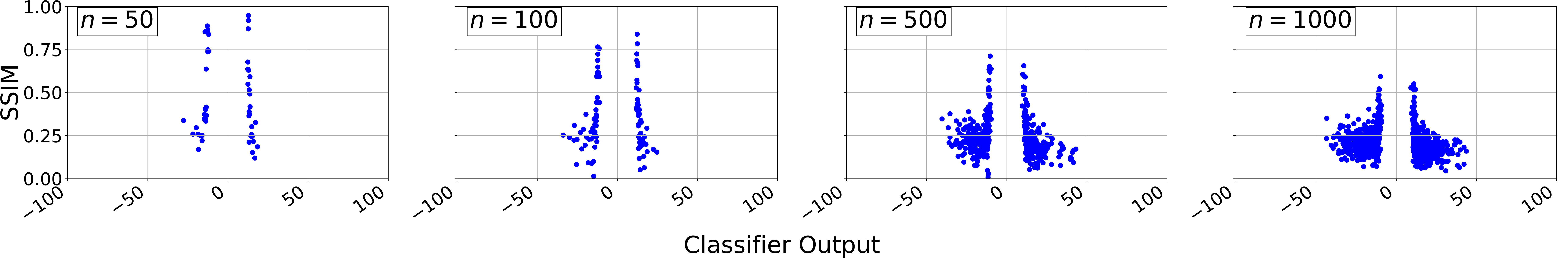} \vspace{6pt}\\
        Models trained on $n=500$ samples with different architectures \\
        \includegraphics[width=\textwidth]{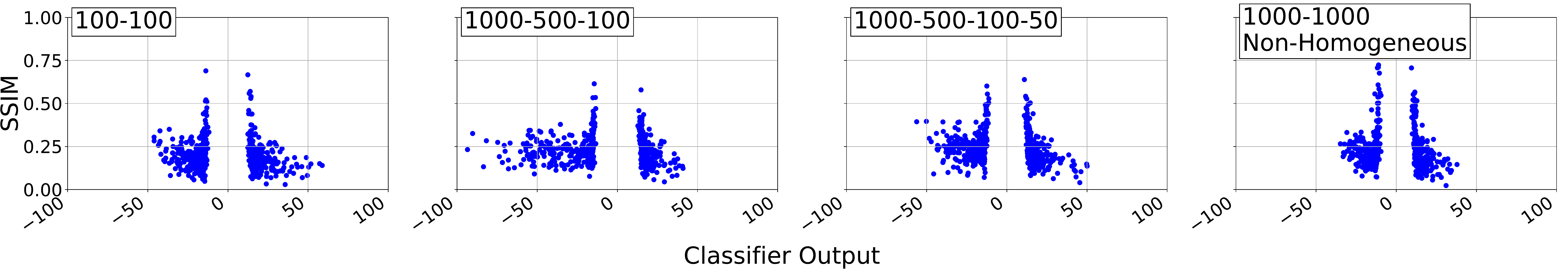} \\
    \end{tabular}
    \caption{Each point represents a training sample. The y-axis is the highest SSIM score achieved by a reconstruction of this sample, the x-axis is the output of the model. 
    \textbf{Top:} The effect of training the same model on different number of training samples ($n$). \textbf{Bottom:} The effect of training models with different architectures (on $n=500$ training samples). The right-most plot shows a $3$-layer non-homogeneous MLP (with bias terms in all hidden layers). See discussion in \secref{subsec:margin}.
    }
    \label{fig:scatter}
\end{figure}

In this section we analyze some relations between our experimental results to the theory laid down in \secref{sec:back and met}. 
Given a trained model and its reconstructed samples, we match each training sample to its best reconstruction (in terms of SSIM score). We then plot this SSIM score against $\Phi(\btheta; \bx)$ (the value of the model's output on this training sample)
-- for all training samples. In \figref{fig:scatter} each cell shows such plot for a given model. The top row shows models trained on the same architecture with different number of training samples ($n$), where in the bottom row we show the results for models trained on $n=500$ training samples, with different architectures (all results are on CIFAR10).

Recall that 
we do not expect to reconstruct samples that are far from the margin~(\subsecref{subsec:dataset_reconstruction}). It is evident from \figref{fig:scatter} that good reconstructions (e.g., SSIM$>0.4$) are obtained for samples that lie on the margin, as expected from theory. The plots indicate that increasing training size makes reconstruction more difficult.
Lastly, as seen from the rightmost plot in the bottom row, we manage to get high-quality reconstructions from a non-homogeneous model (trained with biases in all hidden layers). This indicates that our approach may work beyond the theoretical limitations of \thmref{thm:known KKT}.

\subsection{Comparison to other Reconstruction Schemes}
\label{subsec:comparison}

\begin{figure}
    \centering
    \includegraphics[width=\textwidth]{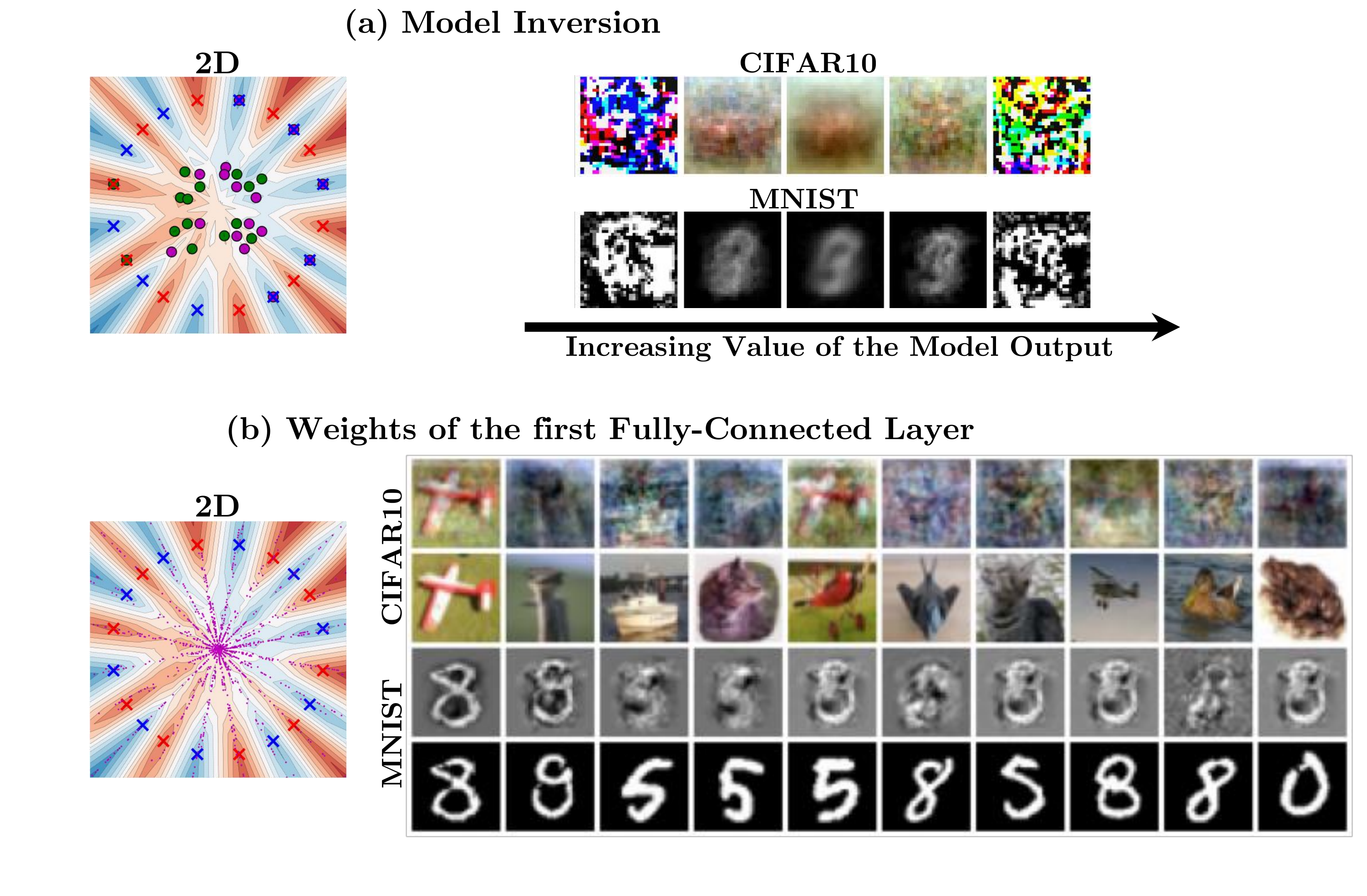}
\vspace*{-1cm}
    \caption{
    Comparison to other reconstruction schemes. \textbf{Top:} Model inversion on the 2D experiment (left), on CIFAR10 (top right) and MNIST (bottom right). The CIFAR and MNIST images are ordered by the value of their output from left (smallest) to right (largest). \textbf{Bottom:} Weights of the first (fully-connected) layer for the 2D experiment (left), CIFAR10 (top right) and MNIST (bottom right). The weights for the 2D experiment are the small purple dots. For the CIFAR and MNIST experiments we show the $10$ weights with the highest SSIM score.
    }
    \label{fig:comparison}
\vspace*{-0.3cm}
\end{figure}

\paragraph{Model Inversion.}
Given a trained model $\Phi(\btheta; \cdot)$, we search for $\bx$ which maximizes or minimizes $\Phi(\btheta; \bx)$, corresponding to positive or negative labels. We initialize 
$\bx \sim \mathcal{N}(0,\sigma I)$ for several values of $\sigma$ and optimize w.r.t. the model output (see Appendix~\ref{appen:implementation} for the choice of hyperparameters).
In~\figref{fig:comparison}a (left) it is apparent that in our two-dimensional experiment, 
model inversion successfully reconstructed $7$ training samples, which indeed lie on a local minimum or maximum. 
However, note that our scheme reconstructs all $20$ samples (\figref{fig:2d circle}). In high dimensions, namely, in MNIST and CIFAR,
while our scheme can reconstruct a large portion of the training set (\figref{fig:reconstruction}a), model inversion converges to noisy/blurry class representatives that correspond to high/low output values (\figref{fig:comparison}a, right). Such results are typical with model inversion 
since
not all class members from the training set are visually similar (see discussions in \cite{shokri2017membership,melis2019exploiting}).

\paragraph{Weights Visualization.}
The weights of the first fully-connected layer have the same dimension as that of the input. One may wonder whether training samples are directly encoded there. In \figref{fig:comparison}b we show the weights that are most similar (SSIM) to a training sample, or all of them in the 2D case. As seen in the 2D case, most weights are in the general direction of a training sample, however the scale is unknown without prior knowledge on the data. For images (MNIST/CIFAR10), not more than $3$ or $4$ of the weights have resemblance to training samples, while our scheme manages to reconstruct dozens of samples. See Appendix~\ref{appen:implementation} and~\ref{appen:results} for details and all $1000$ weights of the models.

\section{Discussion and Conclusion}
\label{sec:challenges and directions}
Even though our results are shown for relatively small-scale models, they are the first to show that the parameters of trained networks may contain enough information to fully reconstruct training samples, and the first to \emph{reconstruct a substantial amount of training samples}.
Moreover, the theoretical basis of the implicit bias in neural networks provides an analytic explanation to this phenomenon.

Solving our optimization problem for convolutional neural networks turned out to be more challenging
and is therefore a subject of future research. We note that the theoretical results that we rely on (i.e., \thmref{thm:known KKT}) also covers convolutional neural networks. We believe that the homogeneity restriction might be relaxed, and showed reconstructions also from a non-homogeneous model (\figref{fig:scatter}, bottom-rightmost). We also believe that our method may be extended to multi-class classifiers using an extension of \thmref{thm:known KKT}. 
Finally, showing reconstructions on larger models and datasets, or on tabular or textual data are interesting future directions.

On the theoretical side, it is not entirely clear why our optimization problem in \eqref{eq:extraction loss} converges to actual training samples, even though there is no guarantee that the solution is unique, especially when using no prior (other than simple bounding to $[-1,1]$). As a final note, our work brings up the question: \emph{are samples on margin the only ones that can be recovered from a trained classifier?} or there exist better reconstruction schemes to reconstruct even more training samples from a trained neural network.

\subsection*{Acknowledgements}
This project received funding from the European Research Council (ERC) under the European Union’s Horizon 2020 research and innovation programme (grant agreement No 788535), and ERC grant 754705, and from the D. Dan and Betty Kahn Foundation, and was supported by the Carolito Stiftung.

% \newpage
\bibliographystyle{abbrvnat}
\bibliography{my_bib}

\newpage

\section*{Checklist}

\begin{enumerate}

\item For all authors...
\begin{enumerate}
  \item Do the main claims made in the abstract and introduction accurately reflect the paper's contributions and scope?
    \answerYes{}
  \item Did you describe the limitations of your work?
    \answerYes{}
  \item Did you discuss any potential negative societal impacts of your work?
    \answerYes{}
  \item Have you read the ethics review guidelines and ensured that your paper conforms to them?
    \answerYes{}
\end{enumerate}

\item If you are including theoretical results...
\begin{enumerate}
  \item Did you state the full set of assumptions of all theoretical results?
    \answerYes{}
        \item Did you include complete proofs of all theoretical results?
    \answerNA{We rely on known theoretical results, so proofs are not required.}
\end{enumerate}

\item If you ran experiments...
\begin{enumerate}
  \item Did you include the code, data, and instructions needed to reproduce the main experimental results (either in the supplemental material or as a URL)?
    \answerYes{}
  \item Did you specify all the training details (e.g., data splits, hyperparameters, how they were chosen)?
    \answerYes{}
        \item Did you report error bars (e.g., with respect to the random seed after running experiments multiple times)?
    \answerNA{}
        \item Did you include the total amount of compute and the type of resources used (e.g., type of GPUs, internal cluster, or cloud provider)?
    \answerYes{}
\end{enumerate}

\item If you are using existing assets (e.g., code, data, models) or curating/releasing new assets...
\begin{enumerate}
  \item If your work uses existing assets, did you cite the creators?
    \answerYes{}
  \item Did you mention the license of the assets?
    \answerYes{}
  \item Did you include any new assets either in the supplemental material or as a URL?
    \answerNA{We do not have new assets.}
  \item Did you discuss whether and how consent was obtained from people whose data you're using/curating?
    \answerNA{We used only publicly available assets.}
  \item Did you discuss whether the data you are using/curating contains personally identifiable information or offensive content?
    \answerNA{We used only publicly available data.}
\end{enumerate}

\item If you used crowdsourcing or conducted research with human subjects...
\begin{enumerate}
  \item Did you include the full text of instructions given to participants and screenshots, if applicable?
    \answerNA{}
  \item Did you describe any potential participant risks, with links to Institutional Review Board (IRB) approvals, if applicable?
    \answerNA{}
  \item Did you include the estimated hourly wage paid to participants and the total amount spent on participant compensation?
    \answerNA{}
\end{enumerate}

\end{enumerate}

\newpage

\appendix

\addcontentsline{toc}{section}{Appendix} % Add the appendix text to the document TOC
\part{Appendix} % Start the appendix part
\parttoc % Insert the appendix TOC
% \newpage

\section{More Details on Privacy Attacks in Deep Learning} \label{app:more related}

Below we discuss several privacy attacks that have been extensively studied in recent years (see \cite{liu2021machine,jegorova2021survey} for surveys).

\paragraph{Membership Inference.}
In \emph{membership-inference attacks} \citep{shokri2017membership,long2018understanding,salem2018ml,yeom2018privacy,song2021systematic} the adversary 
determines whether a given data point was used to train the model or not.
For example, if the model was trained on records of patients with a certain disease, the adversary might learn that an individual's record appeared in the training set and thus infer that the owner of the record has the disease with high chance. Note that membership inference attacks are significantly different from our attack, as the adversary must choose a specific data point. E.g., if the inputs are images, then the adversary must be able to guess a specific image.

\paragraph{Model Extraction.} 
In \emph{model-extraction attacks} \citep{tramer2016stealing,oh2019towards,wang2018stealing,carlini2020cryptanalytic,jagielski2020high,milli2019model,rolnick2020reverse,chen2021efficiently} the adversary aims to steal the trained model functionality. In this attack, the adversary only has black-box access with no prior knowledge of the model parameters or training data, and the outcome of the attack is a model that is approximately the same as the target model. It was shown that in certain cases the adversary can reconstruct the exact parameters of the target model. We note that such attacks might be combined with our attack in order to allow extraction of the training dataset in a black-box setting. Namely, in the first stage the model is extracted using model-extraction attacks, and in the second stage the training dataset is reconstructed using our attack.

\paragraph{Model Inversion.}
\emph{Model-inversion attacks} \citep{fredrikson2015model} are perhaps the closest to our attack, as they consider reconstruction of input data given a trained model. These attacks aim to infer class features or construct class representatives, given that the adversary has some access (either black-box or white-box) to a model.

\cite{fredrikson2015model} showed that a face-recognition model can be used to reconstruct images of a certain person. This is done by using gradient descent for obtaining an input that maximizes the output probability that the face-recognition model assigns to a specific class. Thus, if a class contains only images of a certain individual, then by maximizing the output probability for this class we obtain an image that might be visually similar to an image of that person. It is important to note that the reconstructed image is not an actual example from the training set. Namely, it is an image that contains features which the classifier identifies with the class, and hence it might be visually similar to any image of the individual (including images from the training set). 
If the class members are not all visually similar (which is generally the case), then the results of model inversion do not look like the training data (see discussions in \cite{shokri2017membership} and \cite{melis2019exploiting}). For example, if this approach is applied to the CIFAR-10 dataset, it results in images which are not human-recognizable \citep{shokri2017membership}.
In \cite{zhang2020secret}, the authors leverage partial public information to learn
a distributional prior via generative adversarial networks (GANs) and use it to guide the inversion process. That is, they generate images where the target model outputs a high probability for the considered class (as in \cite{fredrikson2015model}), but also encourage realistic images using GAN. We emphasize that from the reasons discussed above, this method does not reconstruct any specific training data point.
Another approach for model inversion is training a model that acts as an inverse of the target model \citep{yang2019neural}. Thus, the inverse model 
takes the predicted confidence vectors of the target model as input, and outputs reconstructed data. A recent paper \cite{balle2022reconstructing} shows a reconstruction attack where the attacker has information about all the data samples except for one. On the theoretical side, \cite{brown2021memorization} prove that in certain settings, models memorize information about training examples, and show reconstruction attacks on some synthetic datasets.

Model inversion and information leakage in \emph{collaborative deep learning} was studied in, e.g., \cite{he2019model,melis2019exploiting,hitaj2017deep,zhu2019deep,yin2021see,huang2021evaluating}. 
Extraction of training data from language models was studied in \cite{carlini2021extracting,carlini2019secret}, where they use the ability of language models to complete a given sentence in order to reveal sensitive information from the training data.
We note that this attack is specific to language models, which are generative models, while our approach considers classifiers and is less specific.

\paragraph{Defences against Training Data Reconstruction.}
Avoiding leakage of sensitive information on the training dataset is the motivation behind \emph{differential privacy} in machine learning, which has been extensively studied in recent years \citep{abadi2016deep,dwork2006calibrating,chaudhuri2011differentially}.
This approach allows provable guarantees on privacy,
but it typically comes with high cost in accuracy. 
Other approaches for protecting the privacy of the training set, which do not allow such provable guarantees, have also been suggested (e.g., \cite{huang2020instahide,carlini2020private}).

\vspace{-8pt}
\section{Implementation Details}
\label{appen:implementation}
\vspace{-8pt}

\subsection{Hardware, Software and Running Time}
A typical reconstruction runs for about $30$ minutes on a GPU Tesla V-100 $32$GB, for reconstructing $m=1000$ samples from a model with architecture $d$-$1000$-$1000$-$1$, and for $100,000$ epochs (running times slightly differ with the number of samples $m$, number of epochs and the size of the model, but it still takes about this time to run). Our code is implemented in \textsc{PyTorch}~\citep{paszke2019pytorch}. We will release the code.

\subsection{Hyperparameters}

Our reconstruction scheme has $4$ hyperparameters. Already discussed in the paper are the learning rate and $\sigma_\bx$ (discussed in \subsecref{subsec:ts_reconstruct}). In \subsecref{subsec:ts_reconstruct} we discuss the modification in the derivative of a ReLU layer ${y = \text{max}\{0,x\}}$.
The backward function of a ReLU layer works as follows: given the ``gradient from above'' $\frac{\partial L}{\partial y}$, the backward gradient is $\frac{\partial L}{\partial y} \cdot \mathbb{I}\{\bx>0\}$. Our modification to the backward gradient is $\frac{\partial L}{\partial y} \cdot \sigma \left(\alpha \bx \right)$, where $\sigma(z)=\frac{1}{1+e^{-z}}$ and $\alpha$ is a hyperparameter. As noted in the paper, this derivative is essentially the derivative of a SoftReLU, where the derivative is the same as ReLU for $\alpha \to \infty$ and is the derivative of the identity function for $\alpha \to 0$. Note that this is done only in the backward function, while the forward function remains that of a ReLU function.
We also add an extra hyperparameter $\lambda_\text{min}$ to our $L_{\lambda}$ loss from~\eqref{eq:lambda loss}:
$$L_\lambda(\lambda_1,\dots,\lambda_m) = \sum_{i=1}^m \max\{-\lambda_i+\lambda_\text{min}, 0\}$$
The intuition behind is to encourage as many samples to lie on a margin, and thus try and reconstruct some sample from the training set.

To sum it all, the hyperparameters of our reconstruction scheme are:
\begin{enumerate}
    \item Reconstruction learning rate
    \item $\sigma_\bx$, the initial scale of $x_i$ initialization
    \item  $\alpha$, of the derivative of the modified ReLU
    \item $\lambda_\text{min}$
\end{enumerate}

To find the set of hyperparamerers we used Weights\&Biases~\citep{wandb}  using a random grid search where the parameters are sampled from the following distributions:
\begin{itemize}
    \item Learning rate, log-uniform in $[10^{-5},1]$
    \item $\sigma_\bx$, log-uniform in $[10^{-6},1]$
    \item ReLU derivative $\alpha$, uniform in $[10,500]$
    \item $\lambda_\text{min}$, log-uniform in $[10^{-4},1]$
\end{itemize}

When searching for hyperparameters for the model inversion results in \subsecref{subsec:comparison} we use the following:
\begin{itemize}
    \item Learning rate, log-uniform in $[10^{-6},1]$
    \item $\sigma_\bx$, log-uniform in $[10^{-7},1]$
\end{itemize}

\subsection{Post-Processing of Reconstructed Samples}
After the reconstruction run ends we want to match the reconstructed samples to samples from the training set. This is done in the following manner:

\begin{enumerate}
    \item \textbf{Scaling.} Each reconstructed sample is stretched to fit into the range $[0,1]$ (by linear transformation of its minimal/maximal values).
    \item \textbf{Searching Nearest Neighbours.} For each training sample from the training set we compute the distance to all reconstructed outputs using NCC~\citep{Lewis95fastnormalized}.
    \item \textbf{Voting.} For each training sample we compute the mean of all the closest nearest neighbours (all reconstructed samples with NCC score largest than $0.9$ of the distance to the closest nearest neighbour). Now we have pairs of trainig-sample and its reconstruction.
    \item \textbf{Sorting.} For each pair we compute its SSIM~\citep{wang2004image}, and sort the results by descending order.
\end{enumerate}

\section{Supplementary Results}
\label{appen:results}

\subsection{Results for Models in \figref{fig:scatter}}
In this subsection we provide more details and experiments on each model presented in \figref{fig:scatter}. In Table \ref{tab:scatter_errors} we show the train loss, test error and test loss of each model from \figref{fig:scatter}. All the models achieved a train accuracy of $100\%$. We note that adding more training samples improves the test accuracy, while adding more layers keeps the test accuracy approximately the same. In Figures \ref{fig:100, 1000-1000}-\ref{fig:500, 1000-1000, non-homo} we show the best $45$ extracted images (sorted by SSIM score) for the models presented in \figref{fig:scatter}. The reconstructions for the $50$ and $500$ samples with a $d$-$1000$-$1000$-$1$ architecture is presented in \figref{fig:teaser} and \figref{fig:reconstruction} (top) respectively.

\begin{table}[ht!]
    \centering
    \begin{tabular}{ll|ccc}
         Architecture & Training Set Size ($n$) & Train Loss & Test Accuracy & Test Loss  \\
         \midrule
         $1000$-$1000$ & $50$ & $1.5\cdot10^{-6}$ & $72\%$ & $2.14$ \\
         $1000$-$1000$ & $100$ & $2.0\cdot10^{-6}$ & $74\%$ & $2.41$ \\
         $1000$-$1000$ & $500$ & $4.0\cdot10^{-6}$ & $78\%$ & $2.09$ \\
         $1000$-$1000$ & $1000$ & $5.5\cdot10^{-6}$ & $79\%$ & $1.96$ \\
         $100$-$100$ & $500$ & $3.0\cdot10^{-7}$ & $77\%$ & $2.72$ \\
         $1000$-$500$-$100$ & $500$ & $1.2\cdot10^{-7}$ & $78\%$ & $3.14$ \\
         $1000$-$500$-$100$-$50$ & $500$ & $8.4\cdot10^{-7}$ & $77\%$ & $2.57$ \\
         \begin{tabular}[l]{@{}l@{}}$1000$-$1000$\\(Non-Homogeneous)\end{tabular} & $500$ & $4.3\cdot10^{-6}$ & $77\%$ & $2.12$ \\
         \bottomrule
    \end{tabular}
    \caption{Train/Test loss and Test Error for models shown in \figref{fig:scatter}}
    \label{tab:scatter_errors}
\end{table}

\begin{figure}
    \centering
    \includegraphics[scale=0.15]{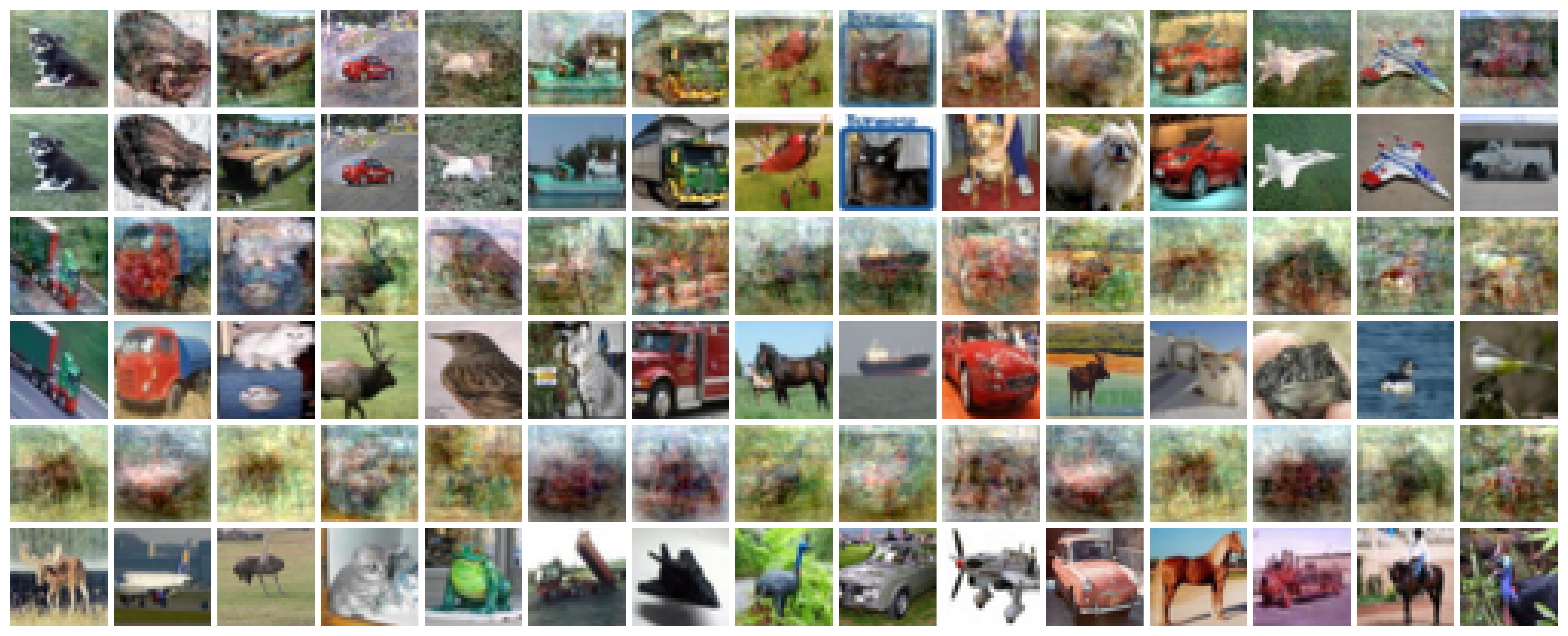}
    \caption{ \textbf{Architecture:} $d$-$1000$-$1000$-$1$, \textbf{Samples:} $100$ \\ Odd rows (1,3,5) are reconstructions, even rows (2,4,6) are the original data.}
    \label{fig:100, 1000-1000}
\end{figure}

\begin{figure}
    \centering
    \includegraphics[scale=0.15]{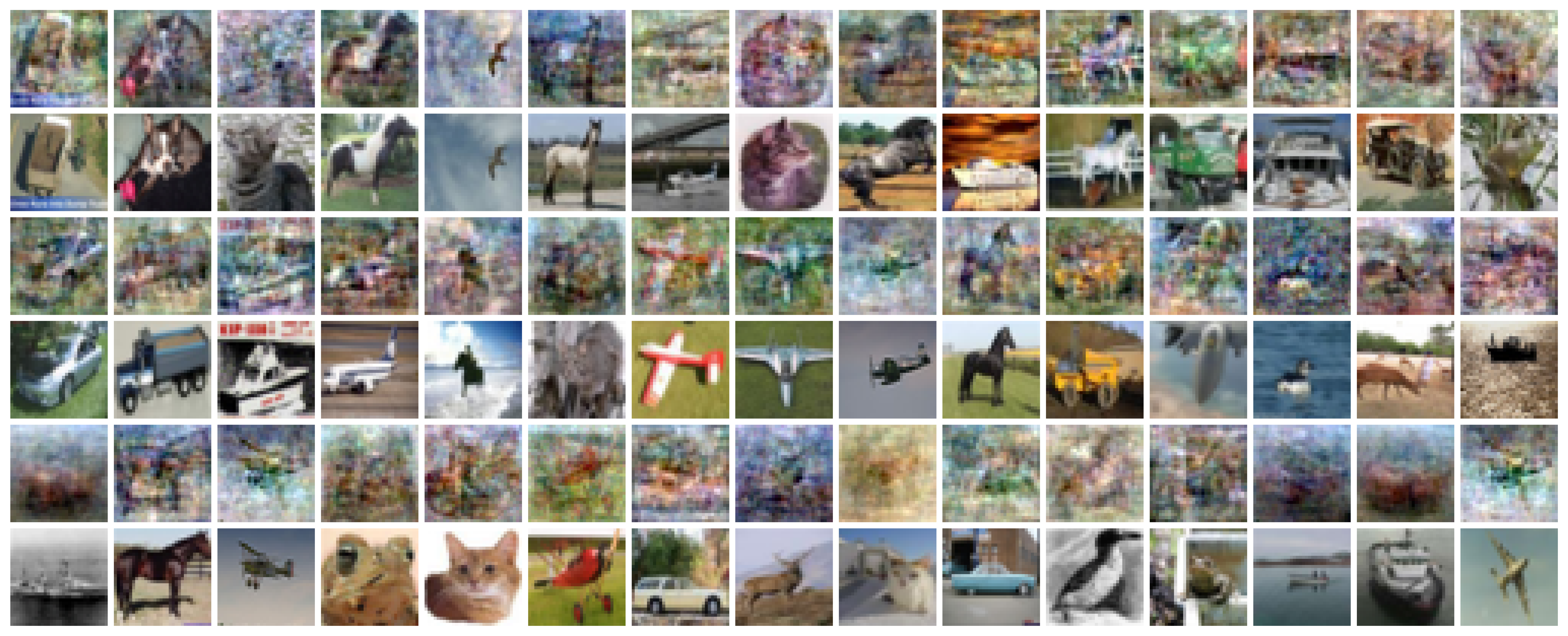}
    \caption{\textbf{Architecture:} $d$-$1000$-$1000$-$1$, \textbf{Samples:} $1000$ \\ Odd rows (1,3,5) are reconstructions, even rows (2,4,6) are the original data.}
\end{figure}

\begin{figure}
    \centering
    \includegraphics[scale=0.15]{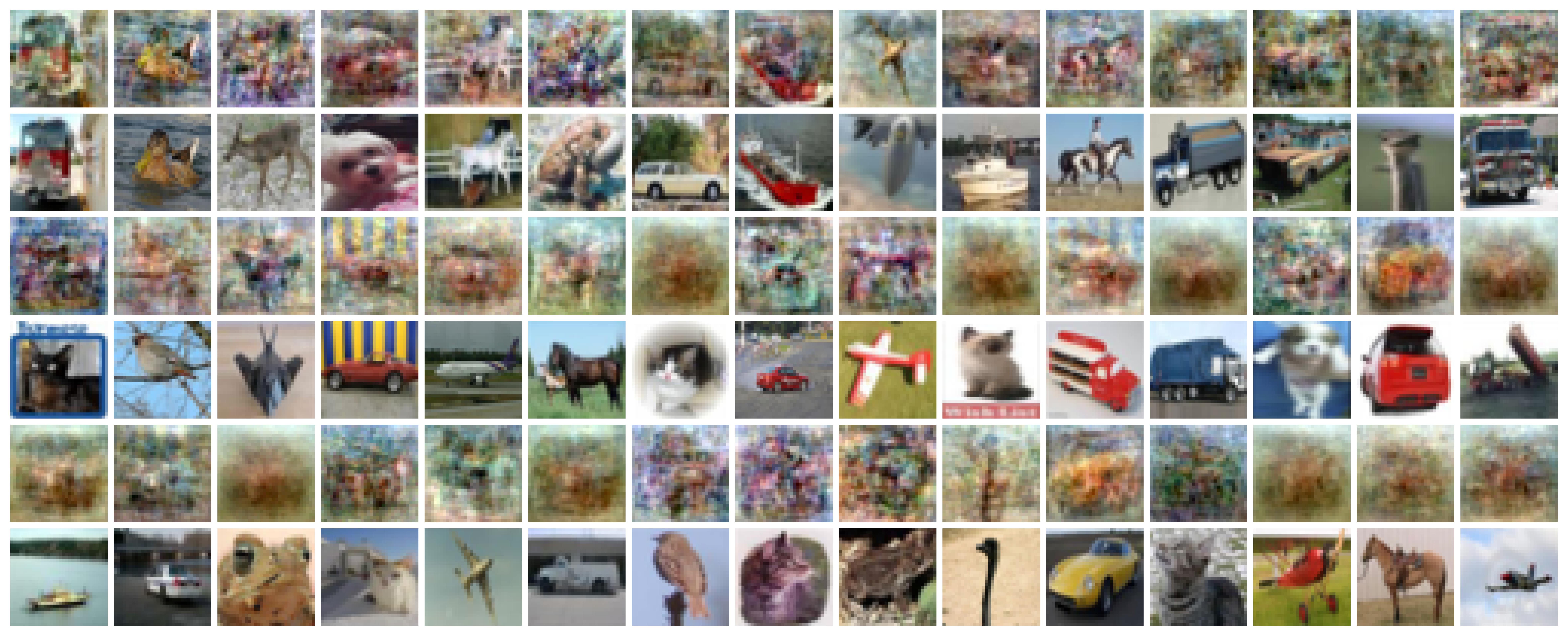}
    \caption{\textbf{Architecture:} $d$-$100$-$100$-$1$, \textbf{Samples:} $500$ \\ Odd rows (1,3,5) are reconstructions, even rows (2,4,6) are the original data.}
\end{figure}

\begin{figure}
    \centering
    \includegraphics[scale=0.15]{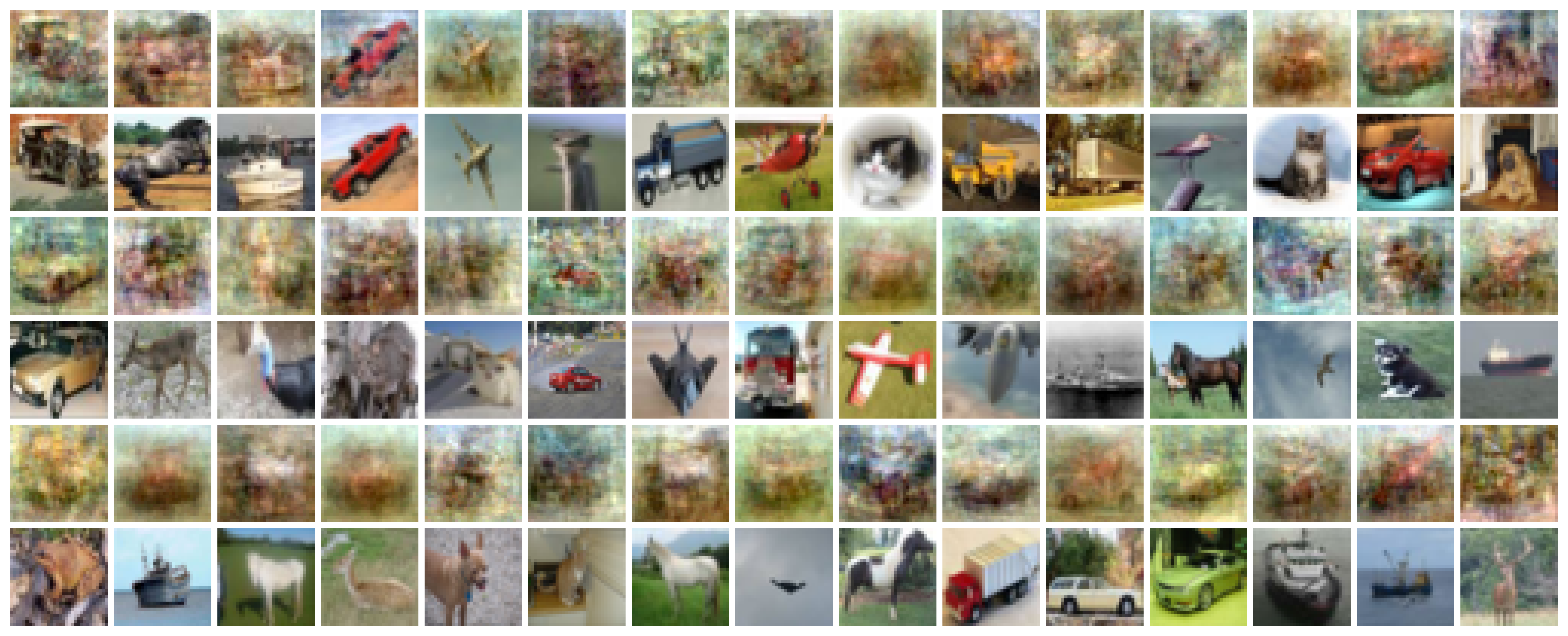}
    \caption{\textbf{Architecture:} $d$-$1000$-$500$-$100$-$1$, \textbf{Samples:} $500$ \\ Odd rows (1,3,5) are reconstructions, even rows (2,4,6) are the original data.}
\end{figure}

\begin{figure}
    \centering
    \includegraphics[scale=0.15]{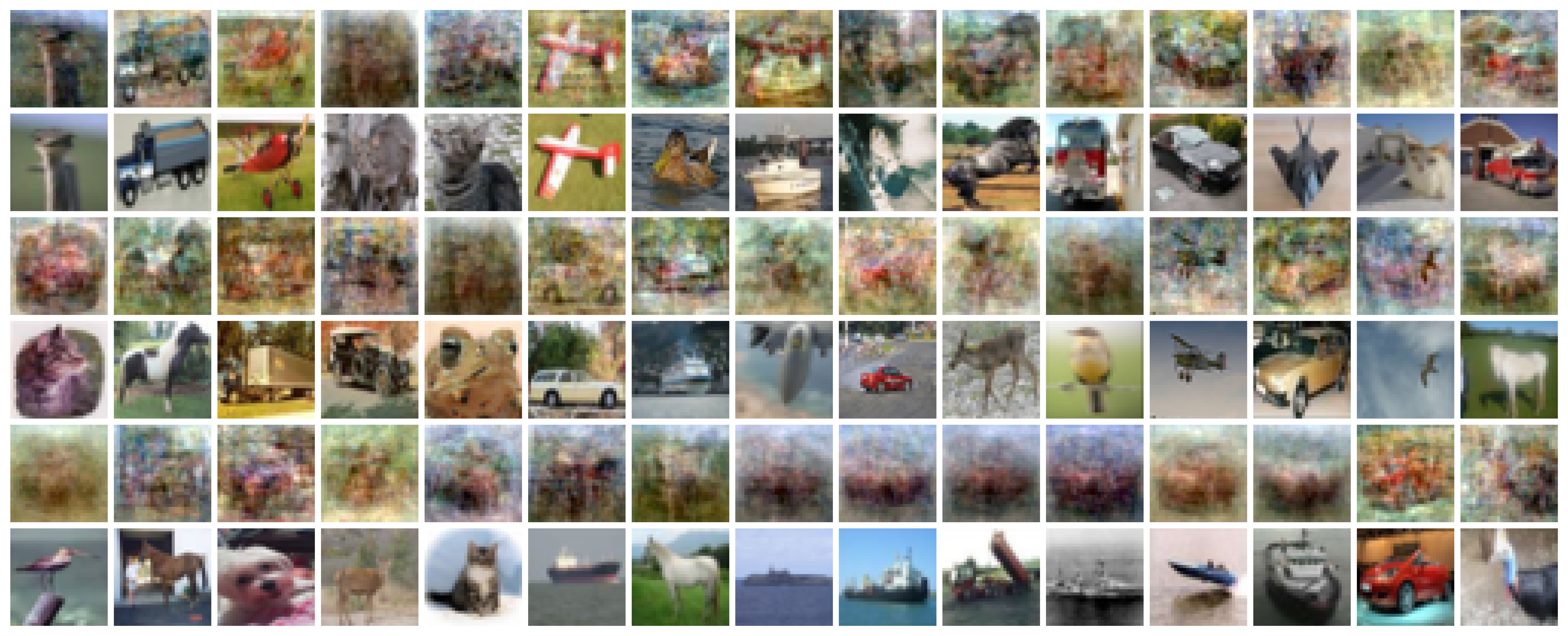}
    \caption{\textbf{Architecture:} $d$-$1000$-$500$-$100$-$50$-$1$, \textbf{Samples:} $500$ \\ Odd rows (1,3,5) are reconstructions, even rows (2,4,6) are the original data.}
\end{figure}

\begin{figure}
    \centering
    \includegraphics[scale=0.15]{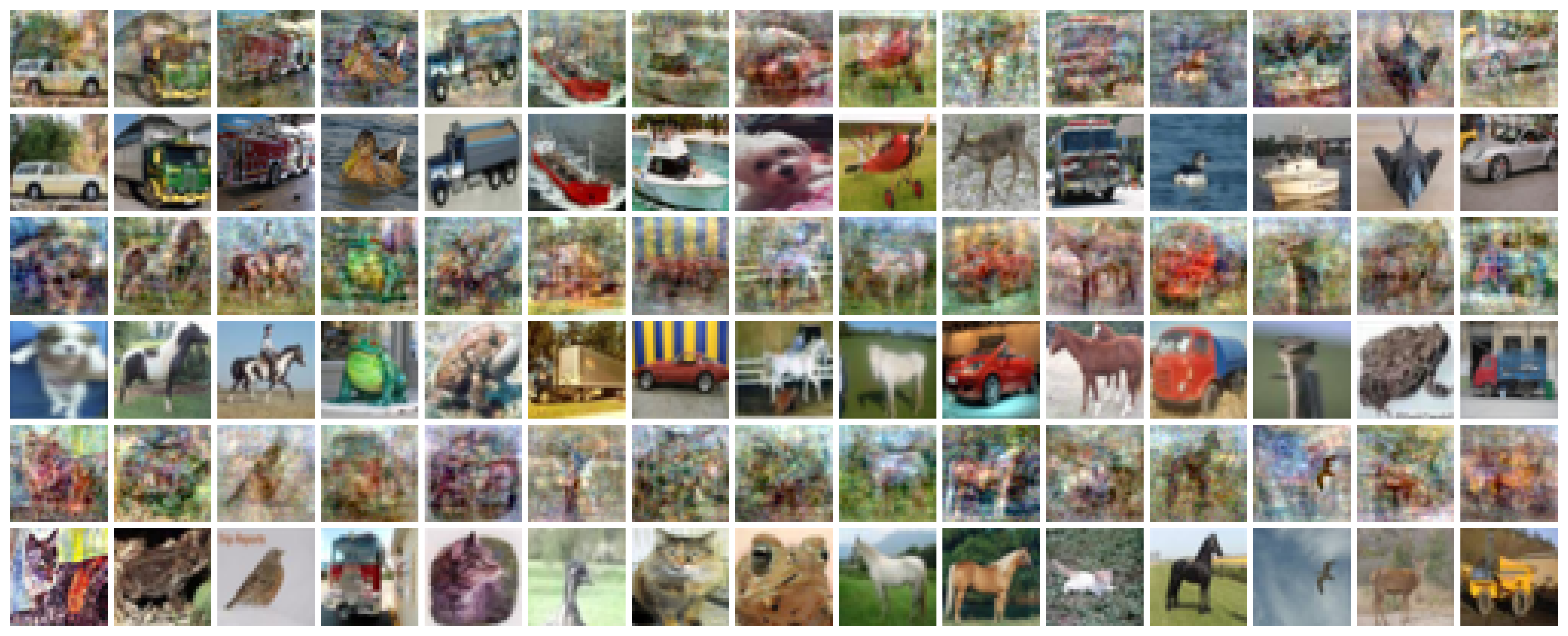}
    \caption{\textbf{Architecture:} $d$-$1000$-$1000$-$1$ (non-homogeneous), \textbf{Samples:} $500$ \\ Odd rows (1,3,5) are reconstructions, even rows (2,4,6) are the original data.}
    \label{fig:500, 1000-1000, non-homo}
\end{figure}

\subsection{All Comparisons for \subsecref{subsec:comparison}}
\begin{figure}
    \centering
    \includegraphics[width=\textwidth]{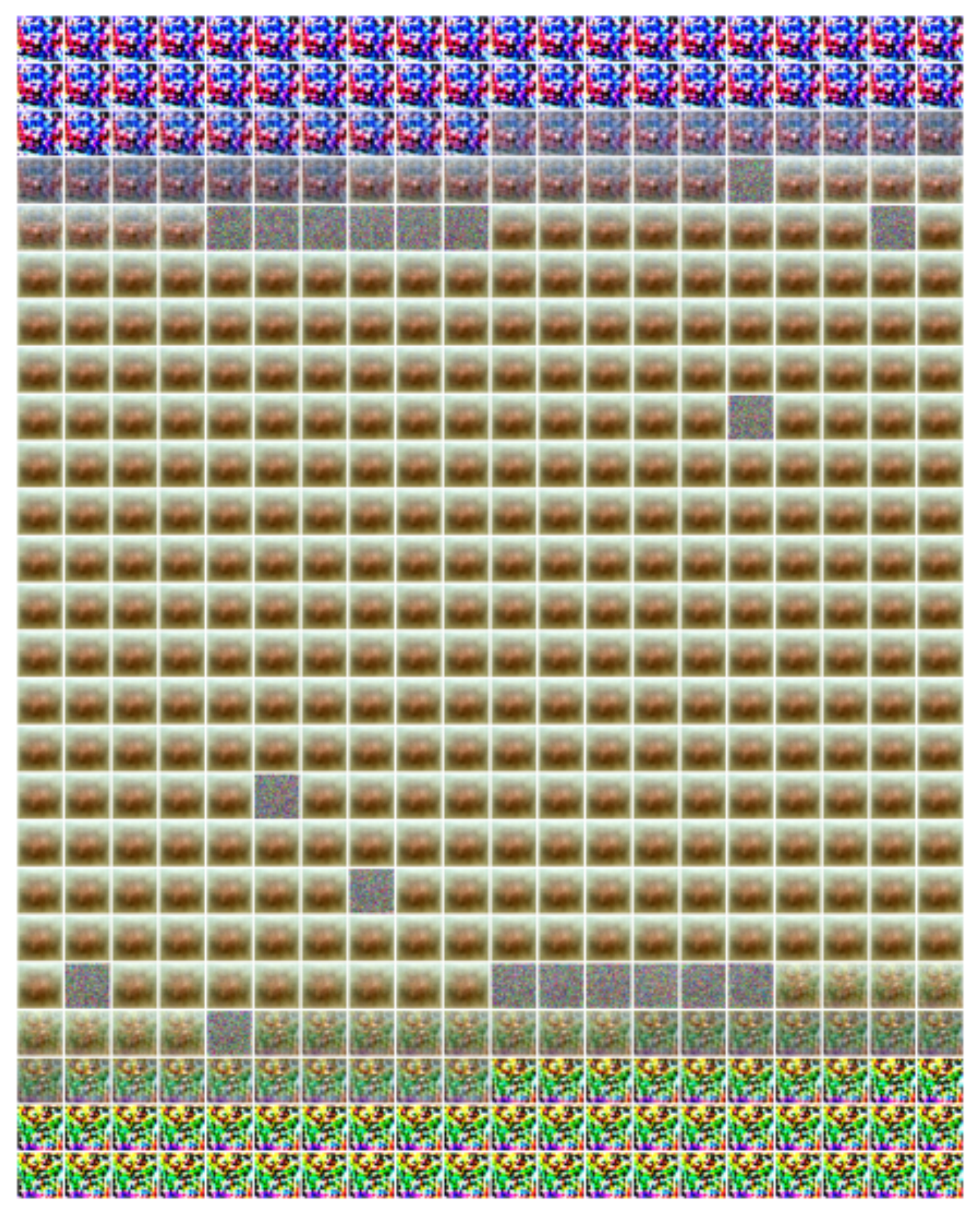}
    \caption{Model inversion attack on a model trained on CIFAR10, with $500$ samples. We reconstructed a total of  $40,000$ images using different initializations and hyperparameters. We sorted the results according to the model's output, and selected $500$ representative with index $i=0,80,160,...,40000$.
    }
    \label{fig:cifar inversion all}
\end{figure}

\begin{figure}
    \centering
    \includegraphics[width=\textwidth]{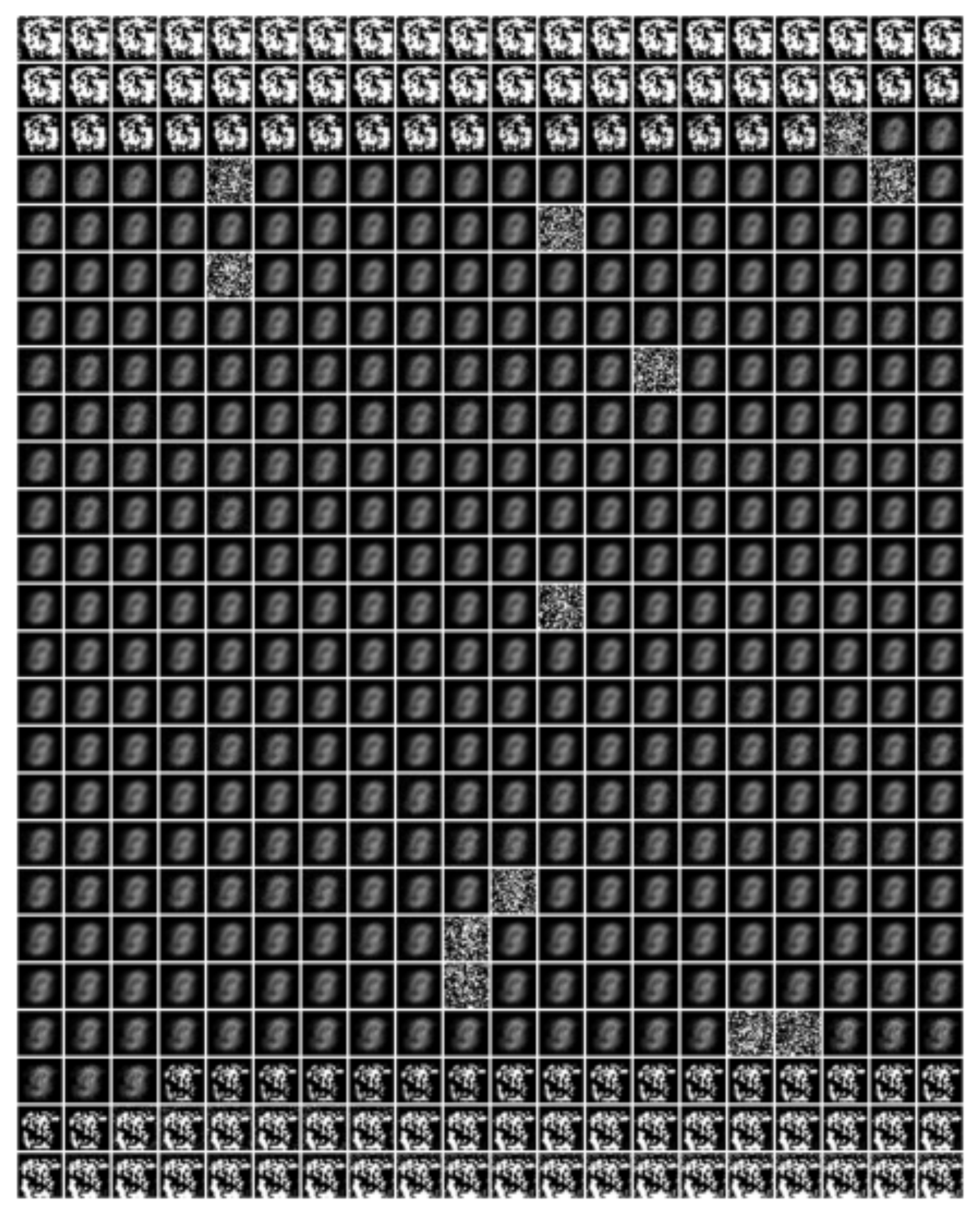}
    \caption{Model inversion attack on a model trained on MNIST, with $500$ samples. We reconstructed a total of  $40,000$ images using different initializations and hyperparameters. We sorted the results according to the model's output, and selected $500$ representative with index $i=0,80,160,...,40000$.
    }
    \label{fig:mnist inversion all}
\end{figure}

In this subsection we provide more detailed results on the comparison to other methods as presented in \subsecref{subsec:comparison}. In \figref{fig:cifar inversion all} and \figref{fig:mnist inversion all} we provide more results from the model inversion attack on models trained on CIFAR10 and MNIST respectively. These are the same models from \figref{fig:comparison} (a). In this attack, we either minimize or maximize the model's output w.r.t. a randomly initialized input. In this experiment, half of the initializations were maximized and the other half is minimized. The images are ordered by output of the model, in an increasing order. The results indicate that the model inversion attack mostly converge to similar reconstructions, even with many different initializations and different hyperparameters. Also, these reconstruction are mostly blurry, and probably represent the averages of each class.

\begin{figure}
    \centering
    \includegraphics[width=\textwidth]{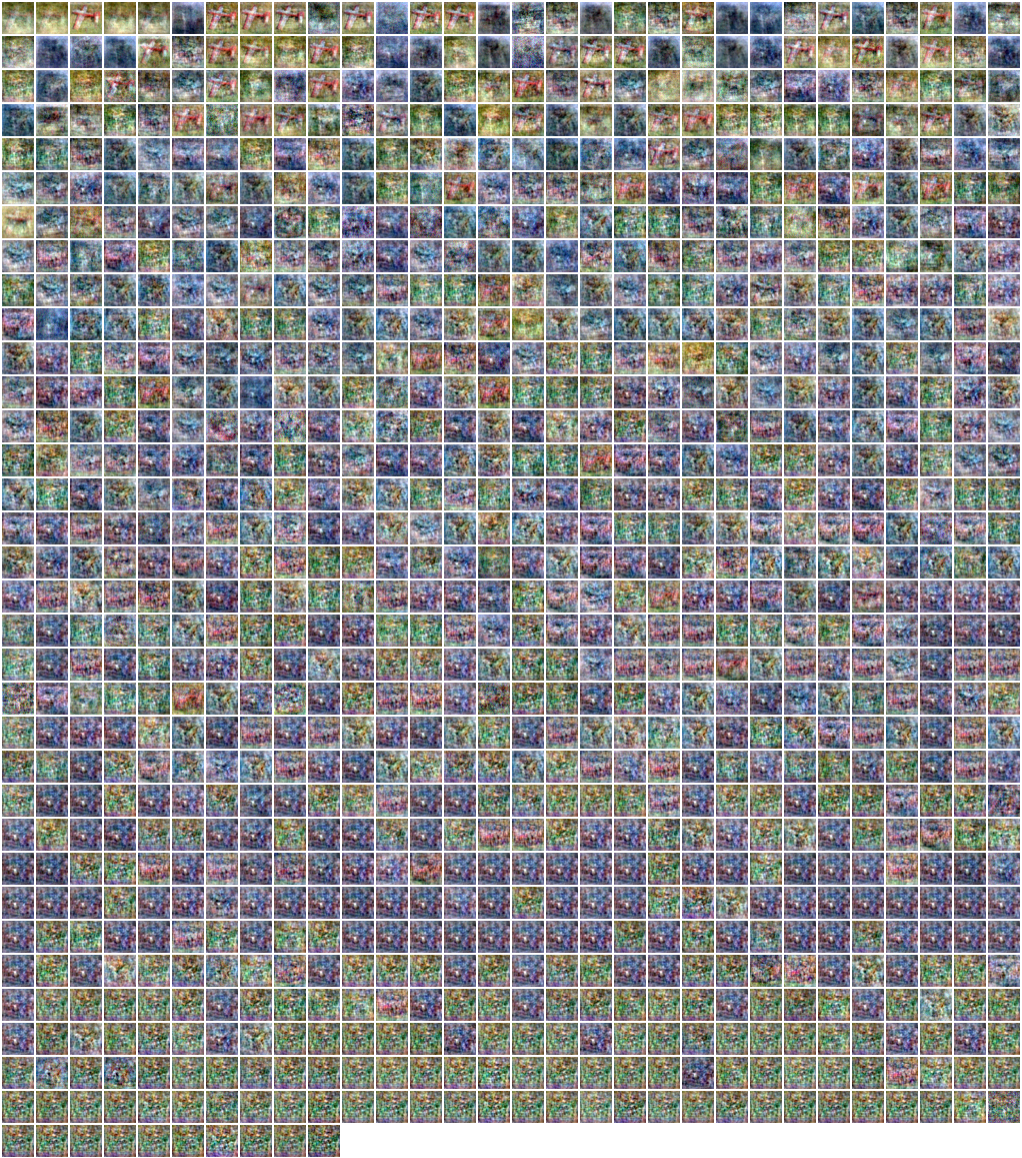}
    \caption{All the $1000$ weights, shown as images, of the first fully-connected layer of a model trained on $500$ samples on CIFAR10.}
    \label{fig:cifar weights all}
\end{figure}

\begin{figure}
    \centering
    \includegraphics[width=\textwidth]{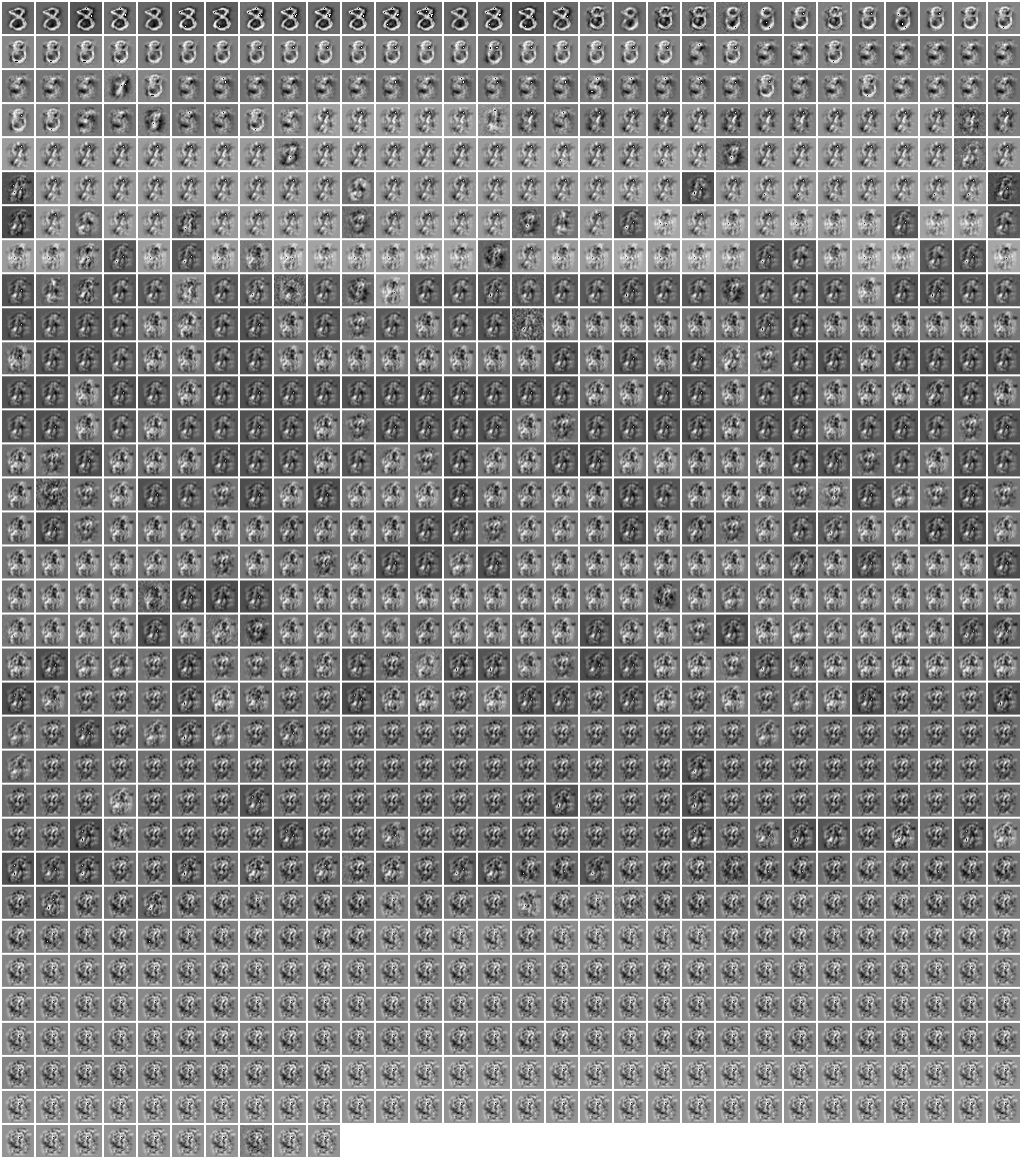}
    \caption{All the $1000$ weights, shown as images, of the first fully-connected layer of a model trained on $500$ samples on MNIST.}
    \label{fig:mnist weights all}
\end{figure}

In \figref{fig:cifar weights all} and \figref{fig:mnist weights all} we show all the weights, as images, of the first fully-connected layer of models trained on CIFAR10 and MNIST respectively. These are the same models as in \figref{fig:reconstruction}, i.e., there are $1000$ weights. Some weights are indicative of several input samples, e.g., a plane from CIFAR10 and the digits $8$ and $5$ from MNIST. We note that our reconstruction scheme is able to reconstruct much more samples, and in better quality than is represented in these weights.

\subsection{Stretching the Theoretical Limitations}
In this section we show results from several experiments which go beyond the theoretical limitations of \thmref{thm:known KKT}.

\begin{figure}[ht!]
    \centering
    \includegraphics[width=\textwidth]{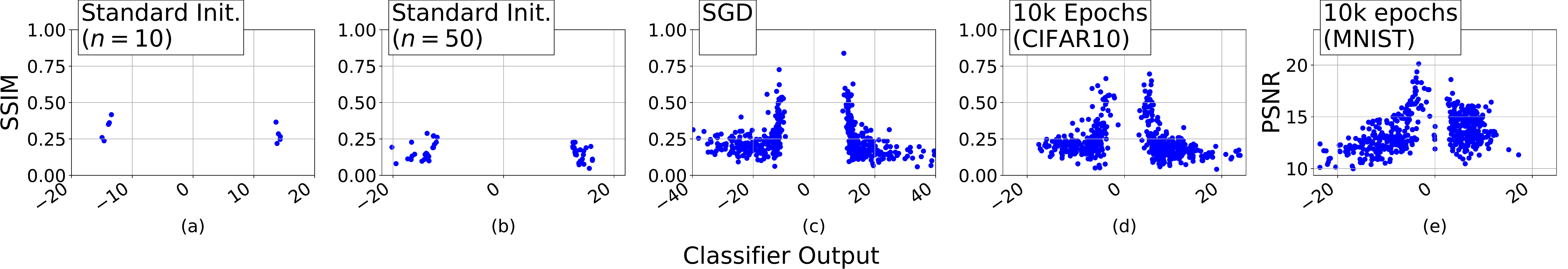}
    \caption{Each point represents a training sample. The y-axis is the highest SSIM score achieved by a reconstruction of this sample, the x-axis is the output of the model. 
    From left to right: (a,b) models trained with standard Kaiming initialization in all layers on $10$ and $50$ CIFAR samples. (c) A model trained using SGD with a batch size of $50$. (d,e) Models trained for $10^4$ epochs on $500$ samples from CIFAR and MNIST respectively.}
    \label{fig:scatter_appendix}
\end{figure}

\begin{table}[ht!]
    \centering
    \begin{tabular}{ll|ccc}
         Experiment & Training Set Size ($n$) & Train Loss & Test Accuracy & Test Loss  \\
         \midrule
         Standard Initialization & $10$  & $8.3\cdot10^{-7}$ & $71\%$ & $1.68$ \\
         Standard Initialization & $50$  & $1.5\cdot10^{-6}$ & $74\%$ & $1.72$ \\
         SGD                     & $500$ & $4.0\cdot10^{-6}$ & $77\%$ & $2.21$ \\
         $10$k Epochs (CIFAR10)  & $500$ & $0.0039$          & $77\%$ & $1.22$ \\
         $10$k Epochs (MNIST)    & $500$ & $0.014$           & $87\%$ & $0.55$ \\
         \bottomrule
    \end{tabular}
    \caption{Train/Test loss and Test Error for models shown in \figref{fig:scatter_appendix}}
    \label{tab:scatter_errors_appendix}
\end{table}

\subsubsection{Standard Initialization Scale}
\begin{figure}[ht!]
    \centering
    \includegraphics[width=\textwidth]{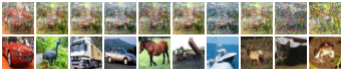}
    \caption{Reconstructions from a model trained on $10$ CIFAR10 images with labels animals vs. vehicles. In the first row are the reconstructions, and in the second row are their corresponding nearest neighbor from the dataset (sorted by SSIM score). }
    \label{fig:10 no init}
\end{figure}

\begin{figure}[ht!]
    \centering
    \includegraphics[width=\textwidth]{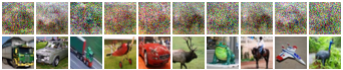}
    \caption{Top $10$ reconstructions from a model trained on $50$ CIFAR10 images with labels animals vs. vehicles. Top row shows reconstructions, and bottom row shows their corresponding nearest neighbor.}
    \label{fig:50 no init}
\end{figure}

In this subsection we consider networks trained with standard initialization scales. We recall that in the experiments presented in \secref{sec:results} the  first fully-connected layer is initialized to a Gaussian distribution with mean $0$ and standard deviation $10^{-4}$, while the other layers are initialized by standard Kaiming initialization \citep{he2015delving}. In \figref{fig:10 no init} and \figref{fig:50 no init} we show reconstructions of a model trained on CIFAR10 on 10 and 50 samples respectively, where the all the layers of the model are initialized by standard Kaiming initialization. The architecture of the model is $d$-$1000$-$1000$-$1$. We note that although the quality of the reconstructions is lower than when initializing the first layer with a small scale, there is still a strong signal that some of reconstructions correlate with training samples. It is an interesting future direction to improve the reconstruction quality for models with standard initialization.

In \figref{fig:scatter_appendix} (a,b) we plot the SSIM score of each training sample against the output of the model. Note that indeed in these experiments the best SSIM score is lower than from other experiments presented in \figref{fig:scatter}. This corresponds to the lower quality of reconstructions when using standard initialization.

\subsubsection{Less Epochs}
\begin{figure}[ht!]
    \centering
    \includegraphics[width=\textwidth]{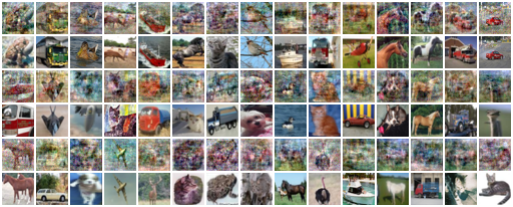}
    \caption{Reconstructions from a model trained for $10^4$ epochs on CIFAR10 with labels animals vs. vehicles. Odd rows (1,3,5) are reconstruction, and even rows (2,4,6) are their nearest neighbor from the training samples.}
    \label{fig:less epochs cifar}
\end{figure}

\begin{figure}[ht!]
    \centering
    \includegraphics[width=\textwidth]{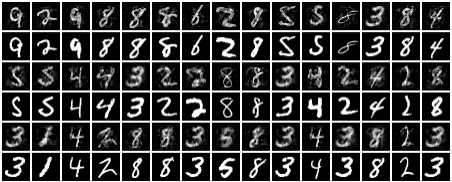}
    \caption{Reconstructions from a model trained for $10^4$ epochs on MNIST with labels odd vs. even. Odd rows (1,3,5) are reconstruction, and even rows (2,4,6) are their nearest neighbor from the training samples.}
    \label{fig:less epochs mnist}
\end{figure}

In the experiments from \secref{sec:results} we trained each model for $10^6$ epochs. The reason for this long training time is that \thmref{thm:known KKT} gives guarantees only when converging to KKT point. Such a convergence happens only after training until infinity, and longer training time may 
%approximately 
converge closer to the KKT point. In this section we provide reconstruction results for models trained for only $10^4$ epochs. \figref{fig:less epochs cifar} and \figref{fig:less epochs mnist} show reconstructions for models trained on $500$ samples from CIFAR10 and MNIST datasets respectively, with an architecture of $d$-$1000$-$1000$-$1$. It is clear that the quality of the reconstruction is very similar to when training for more epochs, this may indicate that even after significantly less training epochs the model converge sufficiently close to a KKT point.

In \figref{fig:scatter_appendix} (d,e) we plot the SSIM score of each training sample against the output of the model. 
We note that we are able to reconstruct samples which appear approximately on the margin for both MNIST and CIFAR. In addition, the model for MNIST did not achieve $0$ train error, and the margin is still very small. With that said, we are still able to reconstruct a large portion of the data with high quality. This goes beyond our theoretical limitations which have guarantees only for models which successfully label the entire training set.

\subsubsection{Mini-batch SGD}
\begin{figure}[ht!]
    \centering
    \includegraphics[width=\textwidth]{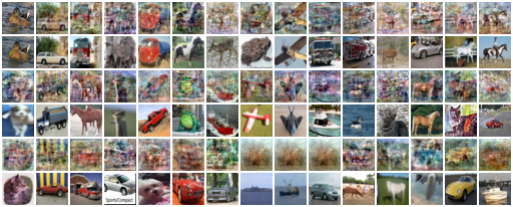}
    \caption{Reconstructions from a model trained using SGD with a batch size of $50$. The model trained on $500$ images from CIFAR10 with labels animals vs. vehicles. Odd rows (1,3,5) are reconstructions and even rows (2,4,6) are their nearest neighbor from the training dataset.}
    \label{fig:sgd}
\end{figure}

In the experiments from \secref{sec:results} we trained the models using full-batch gradient descent. This was done to align with the theoretical guarantees of \thmref{thm:known KKT}, which assume training with gradient flow. In \figref{fig:sgd} we show reconstructions from a model trained with mini-batch SGD, using a batch size of $50$. The model is trained on $500$ images from CIFAR10, and with an architecture of $d$-$1000$-$1000$-$1$. 

In \figref{fig:scatter_appendix} (c) we plot the SSIM score of each training sample against the output of the model. This plot shows that we indeed reconstruct samples that lie on the margin.

\end{document}